\documentclass{article}

\PassOptionsToPackage{numbers}{natbib}
\usepackage[preprint]{neurips_2026}

\usepackage[utf8]{inputenc} 
\usepackage[T1]{fontenc}    
\usepackage{hyperref}       
\usepackage{url}            
\usepackage{booktabs}       
\usepackage{amsfonts}       
\usepackage{nicefrac}       
\usepackage{microtype}      
\usepackage{xcolor}         
\usepackage{enumitem}
\usepackage[pdftex]{graphicx}
\usepackage{multirow}
\usepackage{rotating}
\usepackage{pdflscape}
\usepackage{longtable}

\title{Algebraic Machine Learning for Small-to-Medium Datasets Is Competitive against Strong Standard Baselines}

\author{
  David M\'endez \\
  Mathematics of Behavior and Intelligence Lab \\
  Champalimaud Foundation, Lisbon \\
  david.mendez@research.fchampalimaud.org \\
 \And
  Fernando Martin-Maroto\\
  Mathematics of Behavior and Intelligence Lab \\
  Champalimaud Foundation, Lisbon \\
 fernando.martinmaroto@neuro.fchampalimaud.org \\
  \And
  Gonzalo G. de Polavieja \\
  Mathematics of Behavior and Intelligence Lab \\
  Champalimaud Foundation, Lisbon \\
 gonzalo.polavieja@neuro.fchampalimaud.org \\
}
\begin{document}
\maketitle
\begin{abstract}
Symbolic methods are generally not considered competitive with strong modern learners on realistic supervised tasks. We evaluate Algebraic Machine Learning (AML), a framework that learns through subdirect decomposition of algebraic structure rather than numerical optimization, against standard baselines on image and tabular classification across varying training-set sizes. We find that AML trained only on training data without using validation or cross-validation outperforms a family of cross-validated baseline methods including CNNs on small to medium image datasets (50--2000 training examples). On tabular datasets in the same size range, XGBoost is overall the best performing method, but AML is nonetheless comparable to methods incorporating task-specific biases such as LightGBM and random forests. AML achieves this competitive performance across two very different types of datasets using a generic algebraic inductive bias, rather than the modality-specific biases built into standard baselines like CNNs for images or XGBoost for tabular data, and requires no cross validation because it has no task-dependent hyperparameters to tune.

\end{abstract}

\section{Introduction}

Supervised learning is usually formulated as numerical optimization over a parameterized model class. In this paradigm, inductive bias is encoded through model architecture, regularization, kernels, or ensemble structure, and performance is often determined not only by the learner but also by hyperparameter selection and validation protocol. This view underlies much of modern machine learning, including neural networks, boosted trees, random forests, and support vector machines~\cite{bishop2006prml,goodfellow2016deep,chen2016xgboost,Breiman2001RandomForests,cortes1995svm}.

Symbolic approaches to artificial intelligence follow a different tradition. Classical systems represented knowledge explicitly, typically through rules, logical relations, or expert-curated structures~\cite{NewellSimon1976,Feigenbaum77,Hayes83}. These systems were successful in domains where knowledge could be formalized, but they depended heavily on expert input and were difficult to maintain or scale. As data became abundant and computational resources improved, the field largely shifted away from symbolic learning toward statistical and parametric methods, and symbolic methods are now in common use mainly in niche scenarios involving explicit reasoning or knowledge representation. A common working assumption is that symbolic methods are not competitive on realistic supervised learning tasks unless they are combined with, or subordinated to, modern numerical learners.

This paper tests that assumption. We study \emph{Algebraic Machine Learning} (AML)~\cite{Maroto,SecondPaperArX,ThirdPaperArX}, a symbolic learning framework in which models are constructed through subdirect decomposition (a way of breaking an algebra into irreducible components; see~\S\ref{sec:background}), rather than through gradient-based or combinatorial optimization of a parameterized objective. AML can incorporate manually specified symbolic relations, but it can also learn directly from data. This makes it possible to evaluate AML not as an expert system, but as a supervised learner.

We focus on classification tasks in two distinct modalities: images and tabular data. Inputs are encoded symbolically from their component features, and the only manually supplied formal knowledge is the order relation between numerical feature values. We provide the necessary background on the algebraic encoding and learning procedure in~\S\ref{sec:background}; the full algebraic framework, theoretical foundations, and the main learning algorithm, Sparse Crossing, are developed in~\cite{Maroto,SecondPaperArX,ThirdPaperArX}.

Our experimental question is whether a generic algebraic inductive bias can compete with standard supervised baselines that use modality-specific biases and cross-validated hyperparameters. To answer this, we compare a single AML run, trained only on the training set, against cross-validated baseline families. The baselines include convolutional neural networks for image classification and boosted-tree methods for tabular classification, giving the standard methods the advantage of task-specific architecture and validation-based model selection. A practical consequence of AML's design is that it does not require a held-out validation set as it does not have task-dependent hyperparameters to tune, which is particularly relevant in low-data regimes where every labeled example is scarce.

Our contributions are as follows:
\begin{enumerate}
\item We provide a systematic empirical evaluation of AML on supervised image and tabular classification across training-set sizes from 50 to 2000 examples.
\item We compare AML against strong cross-validated baselines, including CNNs for images and XGBoost, LightGBM, and random forests for tabular data.
\item We show that AML is highly competitive in low- and medium-data regimes despite using a generic algebraic inductive bias and no task-dependent hyperparameter tuning.
\end{enumerate}
Across twelve standard image datasets, AML with a logistic regression readout is the best-performing method in aggregate over the evaluated training sizes, statistically distinguishable from each baseline; it is also the best-performing method on a plurality of datasets in the 200--1000 example range. Across 29 tabular datasets, AML is not the strongest method overall: XGBoost performs best in aggregate, and is the only baseline statistically distinguishable from AML. AML is otherwise comparable to strong tabular baselines such as LightGBM and random forests, and achieves the best result on a plurality of datasets at 1000 training examples.
These findings suggest that symbolic learning can be competitive in realistic supervised tasks when the symbolic structure is learned from data rather than manually engineered. 

\section{Related Work}
\label{sec:related}

\paragraph{Parameterized learners.}
CNNs encode locality and translation equivariance for vision~\cite{lecun1998gradient}. Boosted-tree methods provide strong priors through additive recursive partitioning~\cite{chen2016xgboost, ke2017lightgbm}. Random forests favor ensembles of shallow trees~\cite{Breiman2001RandomForests}. SVMs favor large-margin separation~\cite{cortes1995svm}. These methods have been the standard tools for supervised classification across the modalities we study, and they share two relevant properties: their inductive biases are tailored to specific input modalities, and their performance depends on hyperparameter selection via cross-validation. We use them as the baselines against which to compare AML's design, which is modality-agnostic and requires no task-dependent hyperparameter tuning.

\paragraph{Low-data and few-shot methods.}
Meta-learning~\cite{finn2017model}, transfer from pretrained representations~\cite{chen2020simple}, and data augmentation address low-data settings by importing external structure. We deliberately exclude these from our comparison: our question is whether algebraic learning is competitive with standard supervised baselines trained from scratch, not whether it beats methods that leverage auxiliary data.

\paragraph{Classical and neurosymbolic AI.}
Classical symbolic AI represents knowledge through rules and logic~\cite{NewellSimon1976, Feigenbaum77, Hayes83}. Neurosymbolic systems~\cite{RiegelEtAl2020,BadreddineEtAl2022} combine symbolic constraints with neural learners, typically using symbolic components to guide or regularize an underlying neural model. AML differs from both: its algebraic structure is learned from data (unlike classical systems) and \emph{is} the model itself (unlike most neurosymbolic approaches).

\paragraph{Algebra in machine learning.}
Algebraic ideas appear in equivariant networks~\cite{CohenWelling2016}, algebraic statistics~\cite{DrtonSturmfelsSullivant2009}, and category-theoretic frameworks~\cite{FongSpivak2018}. In these settings algebra describes model classes or invariances. AML differs in that subdirect decomposition is itself the learning mechanism.

\paragraph{AML foundations.}
The algebraic framework, including mathematical rule recovery guarantees, generalization analysis, and Sparse Crossing algorithm, was developed in~\cite{Maroto,SecondPaperArX,ThirdPaperArX}. A more in-depth introduction to the computational and statistical aspects of learning in AML can be found in \cite{arxiv2025}. The present paper does not modify the AML framework; we use it as developed in this prior work and focus instead on a systematic empirical evaluation against modern baselines.

\section{Background: Algebraic Machine Learning}
\label{sec:background}

We summarize the AML framework developed in~\cite{Maroto,SecondPaperArX,ThirdPaperArX}; the reader is referred to those papers for formal definitions, proofs, and algorithmic details.

In AML, rules and data are encoded as relationships between elements in an algebra, where the word algebra is used as in the language of Universal Algebra. Namely, an algebra is a set \(A\) together with one or several operations \(f : A^n \to A\)~\cite{Burris}. The algebras used in this framework are \emph{semilattices}, where a semilattice is an algebra \(A\) with a single binary operation \(\odot\colon A\times A \to A\), which must additionally be commutative (\(a \odot b = b\odot a\)), associative (\(a \odot (b \odot c) = (a \odot b) \odot c\)) and idempotent (\(a\odot a = a\))~\cite{Davey_Priestley_2002}. Every semilattice carries a natural partial order: \(a\le b\) if and only if \(b = a\odot b\), written equivalently as \emph{\(a\) is in \(b\)}. We use this order relation as the central object throughout learning.

\subsection{Encoding a classification problem into the algebras}\label{subsection:embedding}

Tasks are encoded in this framework by building semilattices tailored to the task at hand. We consider a set \(C\) of \emph{constants}, which are the basic symbols used to represent the data and rules we need to encode a task (a grayscale intensity at a pixel, a value of a categorical or numerical variable, a label in a classification task). A data-point is encoded as the product \(\odot\) of all constants representing its features; we call any such product a \emph{term}.

For example, a grayscale image is encoded by constants \(c_{i,j,s}\), one per pixel position \((i,j)\) and intensity \(s\), and its term is the product \(\odot_{i,j} c_{i,j,s(i,j)}\), where \(s(i,j)\) is the intensity at \((i,j)\).
Tabular data-points are encoded analogously, with one constant per (variable, value) pair; missing values are handled by simply omitting the corresponding constants from the term.

In the learning process, we start with an algebra with no relations between constants, the \emph{freest semilattice}. After training, we obtain an algebra in which each example's term \(t\) satisfies \(c_{l_i} \leq t\) for its true label \(l_i\) (\((c_{l_i}, t)\) is a \emph{positive duple}) and \(c_{l_j}\not\leq t\) for any other label (\((c_{l_j}, t)\) is a \emph{negative duple}), where \(c_l\) is the constant encoding the label \(l\). Order relations between numerical-variable values are encoded similarly via order between their constants; we use two oppositely-ordered chains per numerical variable to capture both ascending and descending information \cite{arxiv2025}. No other formal knowledge is supplied to the system in this paper.

\subsection{Learning by subdirect decomposition}

Every algebra admits a subdirect decomposition into irreducible components~\cite{Birkhoff}. AML tracks this decomposition computationally via \emph{atoms}: elements determined by subsets of constants in \(C\) that, collectively, determine the order relation in the algebra and correspond bijectively to the irreducible components of a subdirect decomposition of the algebra~\cite[Theorem 37]{SecondPaperArX}. An atom \(\phi\) discriminates a duple \(t_1, t_2\) if \(\phi < t_1\) (one of the constants that determine \(\phi\) is in \(t_1\)) and \(\phi\not< t_2\). The inequality \(t_1\le t_2\) holds in the algebra if and only if no atoms discriminate \((t_1, t_2)\).

Learning in AML is thus reduced to finding an atomization satisfying the duples that encode the task: no atom discriminates the positive duples, and at least one atom discriminates each negative duple. The Sparse Crossing algorithm~\cite{Maroto} performs this construction iteratively. Given a current atomization and a batch of duples, it removes atoms discriminating the positive duples and adds atoms to discriminate the negative duples, in a way that generalizes to unseen data.

\subsection{The computational learning pipeline}\label{section:aml_computations}

Our learning pipeline mostly follows that of \cite{arxiv2025}, where it is explained in more detail; we modified the batch sizes to match our smaller number of train examples. In short, we start with the freest semilattice and successively use sparse crossing to enforce batches of duples. In each batch, sparse crossing produces a new model, the \emph{master model}; as well as a model that aggregates information across batches, the \emph{union model}.

We train (i.e., perform sparse crossing) for a maximum of \(1000\) batches. We start with batches consisting of \(1/3\) of all positive and negative example-related duples, and linearly grow the batch size until each batch contains all of the task's duples after \(666\) batches, or \(2/3\) of the maximum number of batches. Duples encoding the order of numerical variables are always enforced. We stop training early if the union model perfectly encodes the entire train dataset for \(40\) consecutive batches.

After training, we reduce the union model to \(10\%\) of its size. We do so in such a way that all atoms are individually discriminative of at least one negative example, collectively discriminate all negative duples in the training set, and are diverse in the duples they discriminate. This both promotes generalization and reduces inference cost. We do so in a loop of independent computations, in each of which, we randomly sort the negative duples, and beginning from an empty set of atoms \(S\), for each duple \(r\), we test if there is already a duple in \(S\) discriminating \(r\) and, if not, randomly sample atoms from the union model until one discriminates \(r\) and add it to \(S\). We perform this iteratively until the union of all of these subsets reaches the desired size.

For test-time evaluation, we cannot directly check which positive duples hold in the model as there are usually examples for which all duples tied to classification are discriminated by some atom. In \cite{arxiv2025} the authors classify in AML in two ways. The first way uses the algebraic information directly: they classify an example as belonging to the class for which the corresponding classification duple is discriminated by the fewest number of atoms, which they call \emph{fewest misses}. The second way uses a \emph{logistic regression readout} on the atomization, thus weighing atoms by their statistical importance for classification. We elected to use the logistic regression readout for our headline results (\S\ref{section:image_classification_results} and \S\ref{section:tabular_classification_results}) as they consistently produce stronger \(F_1\) scores, but we nonetheless evaluate the fewest misses variant in \S\ref{section:logistic_regression_role} to characterize how much of AML's competitiveness depends on the readout versus the algebraic atomization itself.

In the logistic regression readout, each example produces a \(\pm 1\) vector indexed by atoms (entry \(+1\) if the atom is below the example's term, \(-1\) otherwise), and a single fully-connected linear layer is trained on top to predict the class. This readout is trained on the same training data used for the algebra until the per-example average train loss falls below \(10^{-7}\); no validation data is used at any step.

\section{Experimental Setup}
\label{sec:setup}

The same encoding framework is used for both modalities for AML, as per~\S\ref{sec:background}: constants represent primitive feature--value pairs and class labels, and examples are encoded as terms. No modality-specific structure (e.g.\ spatial adjacency for pixels, feature interactions for tabular data) is introduced.

\paragraph{Images.}

Due to AML's computational cost on high-resolution inputs, we restrict to eleven datasets at \(32 \times 32\) or lower resolution, plus COIL-20 at \(128 \times 128\). The selection covers handwriting recognition  (MNIST~\cite{lecun2010mnist}, Kuzushiji-49~\cite{clanuwat2018deep}), medical imaging (BloodMNIST \cite{bloodmnist}, OrganCMNIST~\cite{organcmnist}, PneumoniaMNIST~\cite{pneumoniamnist}, DermaMNIST~\cite{dermamnist1, dermamnist2}, part of MedMNIST~\cite{yang2023medmnist}) and general object recognition (CIFAR-10~\cite{Kri09}, Fashion-MNIST~\cite{xiao17}, STL-10~\cite{Coates2011AnAO}, Aerial Cactus Identification~\cite{aerial-cactus-identification, CactusAerial}, Street View House Number~\cite{Netzer2011ReadingDI}, COIL-20~\cite{CAVE_0188}). Datasets span 2--49 classes, and include both grayscale and color images. STL-10 is down-scaled to \(32 \times 32\) as per~\cite{Coates2011AnAO}; all other datasets are presented at native resolution. We use each dataset's provided train/test split where available; for COIL-20, which has no canonical split, we randomly hold out 22 images per object (440 total) for the test set. Both AML and the baselines receive identical training and test data: the first \(n\) training examples (after shuffling, for COIL-20) are used for each \(n \in \{50, 100, 200, 500, 1000, 2000\}\) (no \(n = 2000\) for COIL-20). Dataset characteristics are detailed in Appendix~\ref{section:datasets_chars}, Table~\ref{table:image_datasets}.

\paragraph{Tabular data.}
We use the 29 tabular datasets selected for primary evaluation of classification tasks in~\cite{Hollmann2025}, which span 2--308 features (mixed categorical and numerical) and 2–10 classes. These datasets do not come with pre-defined splits, so we randomly hold out 10\% of each dataset for testing. For each \(n \in \{50, 100, 200, 500, 1000, 2000\}\), the first \(n\) examples (after shuffling) of the remaining data form the training set, ensuring AML and baselines see identical data. Dataset characteristics are in Appendix~\ref{section:datasets_chars}, Table~\ref{table:tabular_datasets}.

\subsection{Baselines}

All baselines use cross-validated hyperparameter selection. AML uses fixed hyperparameters, with no dataset-specific tuning and no validation data. These parameters include those already introduced in~\S\ref{section:aml_computations}, and other hyperparameters discussed in~\cite{arxiv2025}, see Appendix \ref{section:aml_methods}.

Our baselines comprise MLPs and SVMs (general-purpose), XGBoost, LightGBM, and random forests (tabular-tailored), and CNNs (vision-tailored). All methods are trained on both modalities, except CNNs which we restrict to images. We train 200 runs per method and dataset, using randomly sampled parameters, and select the parameters for the final evaluation using 5-fold cross-validation. The full methodology for the baseline computations can be found in Appendix~\ref{section:baselines_methods}.

\subsection{Evaluation}\label{subsection:eval}

For each dataset, we evaluate the model obtained after a single run of AML following~\S\ref{sec:background}. Our headline results (\S\ref{section:image_classification_results} and \S\ref{section:tabular_classification_results}) make use of the logistic regression readout, and \S\ref{section:logistic_regression_role} compares these results with those of fewest misses empirically. For the baseline methods, the reported scores correspond to refitting to the entire train set a model with the hyperparameters selected in cross-validation. We also fit models with all the sampled hyperparameters, with the purpose of analyzing the distributions of the results. We analyze results in terms of macro-\(F_1\) to handle class imbalance; we report macro-\(F_1\) and accuracy results for every dataset and train-set size, for AML in both classification variants and for all baselines, in Appendix \ref{section:extended_results}.

For statistical significance analysis, we first apply the Friedman test \cite{Demsar06} to test whether method ranks across datasets come from the same distribution. Following \cite{Benavoli16}, we replace the conventional Nemenyi post-hoc test with pairwise two-sided Wilcoxon signed-rank tests, applying Holm's step-down correction \cite{Holm79} over all \({k}\choose {2}\) pairwise comparisons among the \(k\) methods, as discussed in  \cite{Demsar06, GarciaetalExtensionDemsar}. All tests use \(\alpha = 0.05\). To quantify effect sizes, we report the Hodges-Lehmann (HL) estimator of paired difference \cite{Hollander2015} (the median of pairwise averages of per-dataset differences, paired with the Wilcoxon test in the inversion-of-tests sense) and its 95\% confidence interval, alongside raw and Holm-adjusted \(p\)-values. We summarize results using critical-difference diagrams \cite{Demsar06}. All tests are computed using R's \texttt{stats} package.

We aggregate across (dataset, training-size) pairs; per-size analyses, which avoid the assumption of independence across sizes, are reported in Appendix~\ref{section:per_size_stats}.

\section{Results}
\label{sec:results}

\subsection{Image classification}\label{section:image_classification_results}

\begin{figure}
    \begin{center}
        \includegraphics[width=0.48\textwidth]{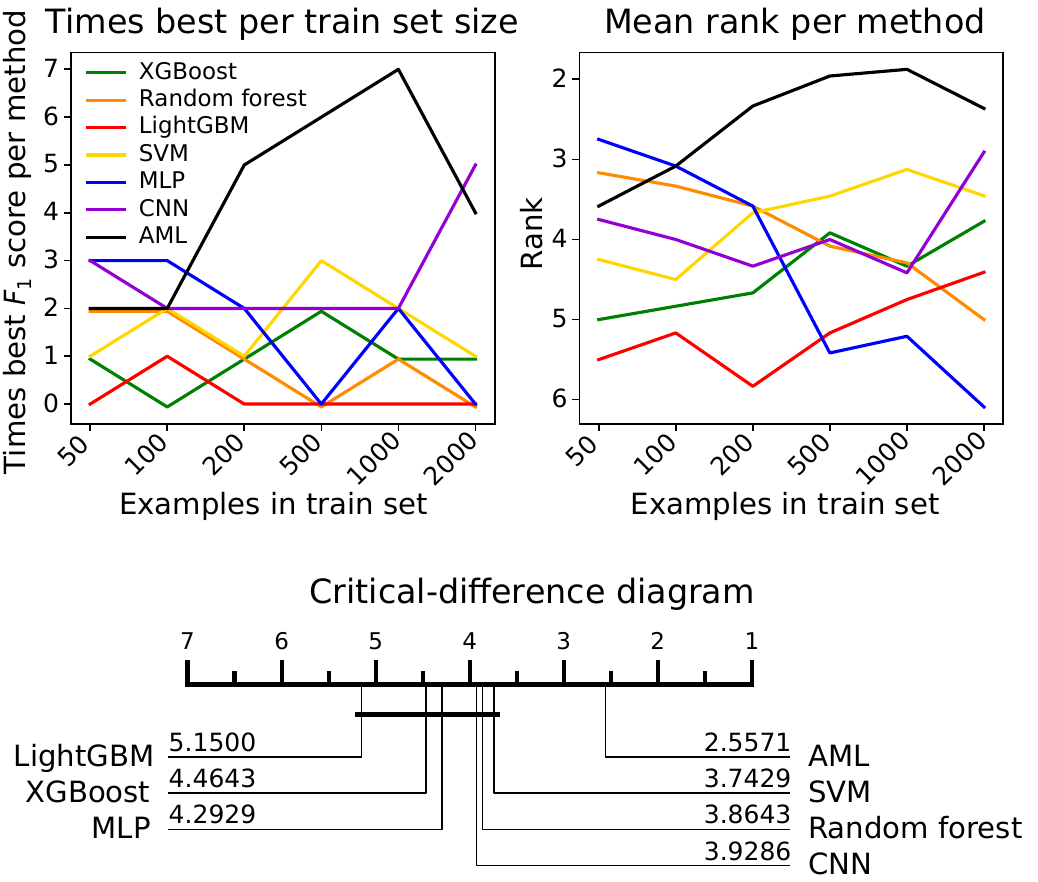}
        \caption{For image datasets, per training-set size, count of how many times each method achieved the best \(F_1\) macro average (top left) and the mean rank of each method (top right); and critical-difference diagram for the aggregate results (bottom).}
        \label{fig:times_best_images}
    \end{center}
\end{figure}

Figure~\ref{fig:times_best_images} shows, for each training-set size, the number of
datasets (out of twelve) on which each method achieves the highest test \(F_1\) macro average (top left), as well as the mean rank of each method at every size (top right). We observe strong results compared to the baselines for train set sizes 200--1000, where AML attains the highest \(F_1\) macro average at least twice as often as every other method. We similarly observe a large gap in the mean ranks. In this regime, standard baselines must simultaneously learn internal representations and select hyperparameters from limited data, while AML constructs algebraic structure directly from the training set with fixed hyperparameters. For the largest considered train-set size (2000 examples), we observe that CNNs become competitive with AML, which is consistent with the expectation that parametrized models exploit their task-specific biases better as data becomes more plentiful. 

When aggregating the data across sizes, the Friedman test rejects the hypothesis that the ranks come from the same distribution (\(p<0.001\)), and the post-hoc test (Table~\ref{table:image_p_values}) further rejects equivalence between AML and each baseline; combined with the positive Hodges-Lehmann estimator in every comparison, this supports AML outperforming each baseline by the magnitudes reported in the table.
Figure~\ref{fig:times_best_images} bottom shows the corresponding critical-difference diagram, in which methods are sorted by mean rank and horizontal bars connect cliques that the post-hoc test cannot statistically distinguish. AML is alone in its clique; the six baselines form a single clique.

\begin{table}
  \caption{Results of post-hoc pairwise Wilcoxon signed-rank tests for AML compared with the baseline methods on image datasets. Positive HL estimator values indicate AML achieves higher \(F_1\) than the baselines.}
  \label{table:image_p_values}
  \centering
  \small
  \begin{tabular}{llllll}
    \toprule
    Method & Raw \(p\)-value & Adjusted \(p\)-value & HL estimator & 95\% CI & Reject \\
    \midrule
        XGBoost & \(<0.001\) & \(<0.001\) & \(+0.0269\) & \([+0.0181, +0.0386]\) & Yes \\[1pt]
        Random forest & \(<0.001\) & \(<0.001\) & \(+0.0157\) & \([+0.0099, +0.0216]\) & Yes \\[1pt]
        LightGBM & \(<0.001\) & \(<0.001\) & \(+0.0334\) & \([+0.0242, +0.0466]\) & Yes \\[1pt]
        SVM & \(0.0020\) & \(0.0274\) & \(+0.0200\) & \([+0.0072, +0.0346]\) & Yes \\[1pt]
        MLP & \(<0.001\) & \(0.0039\)& \(+0.0269\) & \([+0.0127, +0.0413]\) & Yes \\[1pt]
        CNN & \(<0.001\) & \(0.0039\) & \(+0.0279\) & \([+0.0125, +0.0503]\) & Yes \\
    \bottomrule
  \end{tabular}
\end{table}

\begin{figure}
    \begin{center}
        \includegraphics[width=0.94\textwidth]{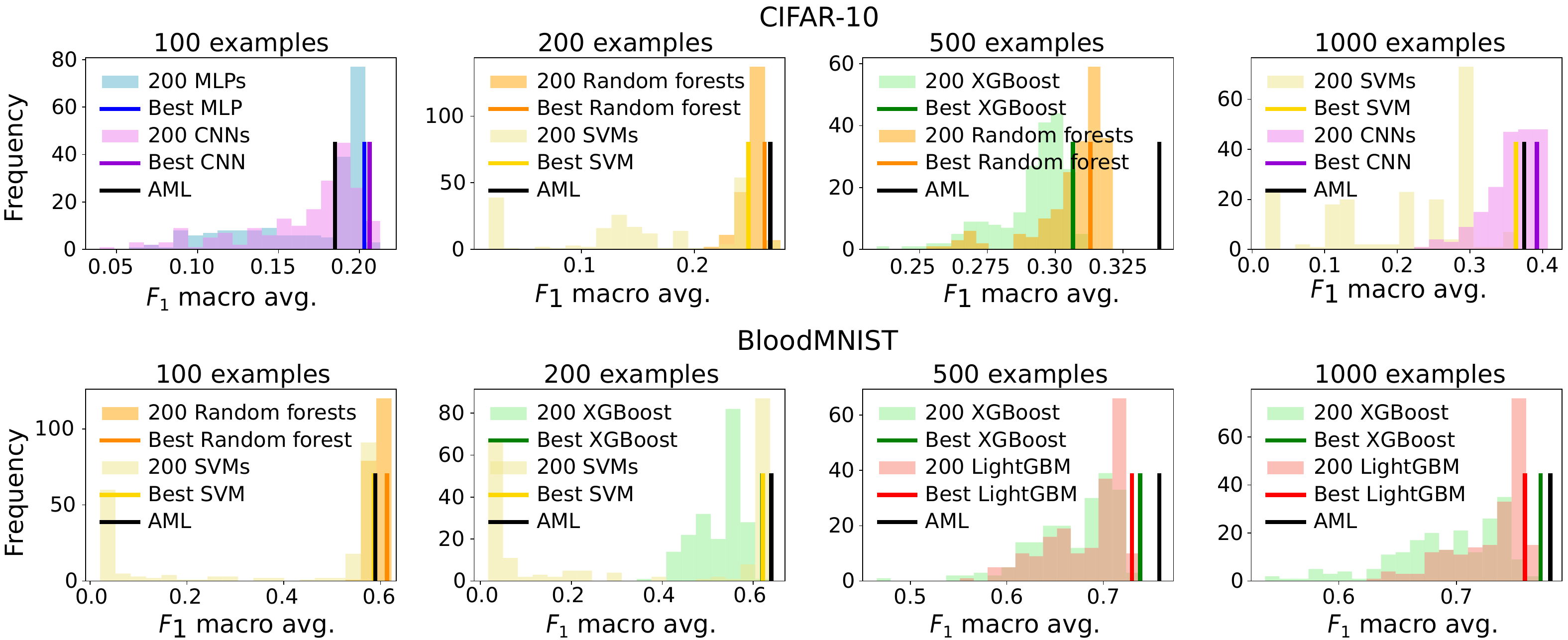}
        \caption{For image datasets, examples of distributions of \(F_1\) scores across all runs of the two baseline models that achieved the best results in cross-validation, highlighting the results of the models trained with the hyperparameters selected in cross-validation (labeled best), and the result for AML, for sizes 100, 200, 500, and 1000.}
        \label{fig:image_distribution}
    \end{center}
\end{figure}

Figure \ref{fig:image_distribution} shows distributions of test \(F_1\) across the 200 baseline training runs for two datasets for 100--1000 examples; the vertical lines mark the AML score and the score of the cross-validated baselines. We generally observe a wide spread of scores amongst the different train runs, particularly for smaller sizes and especially for SVMs. Nonetheless, the parameters chosen by cross-validation generally correspond to runs near the best test scores, especially for larger train-set sizes.

\subsection{Tabular classification}\label{section:tabular_classification_results}

\begin{figure}
    \begin{center}
        \includegraphics[width=0.48\textwidth]{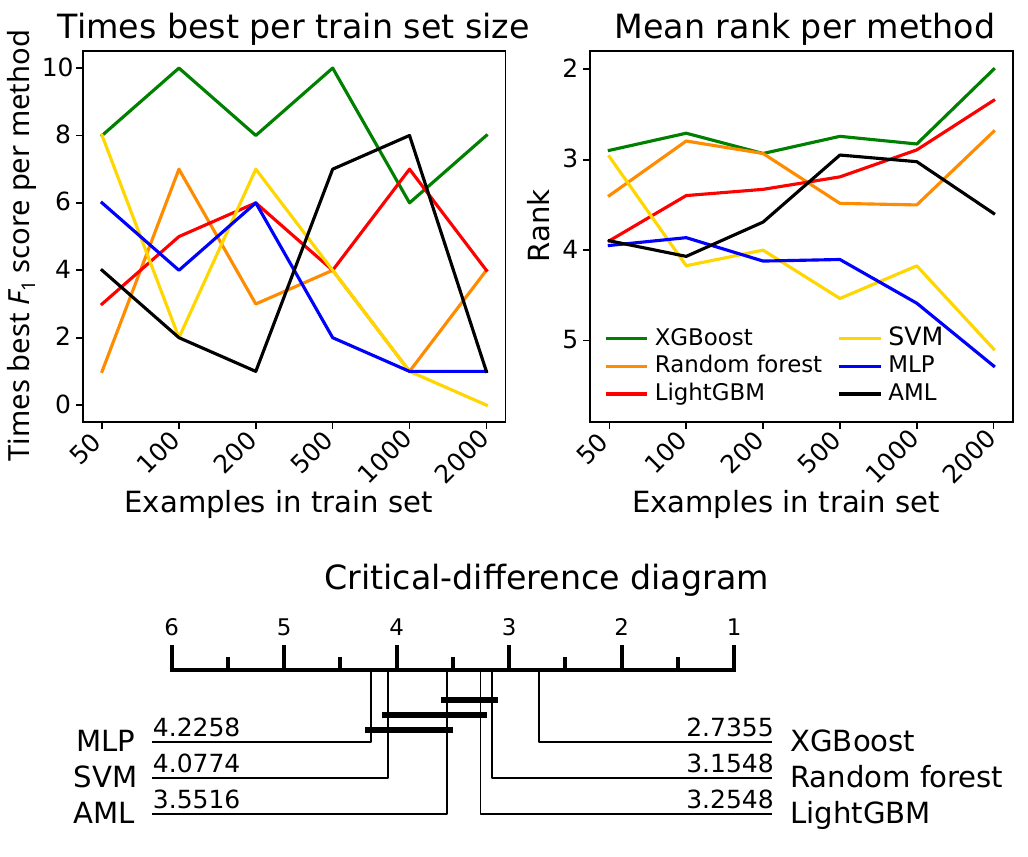}
        \caption{For tabular datasets, per training-set size, count of how many times each method achieved the best \(F_1\) macro average (top left) and the mean rank of each method (top right); and critical-difference diagram of the aggregate results (bottom).}
        \label{fig:times_best_tabular}
    \end{center}
\end{figure}

On tabular tasks, AML's mean rank is competitive with LightGBM and random forests and is lower than that of MLPs and SVMs at sizes greater than 200. This is noteworthy: gradient-boosted trees are widely considered the strongest standard methods for tabular prediction, and AML achieves comparable rank without data-specific biases or hyperparameter tuning.

\begin{figure}
    \begin{center}
        \includegraphics[width=0.94\textwidth]{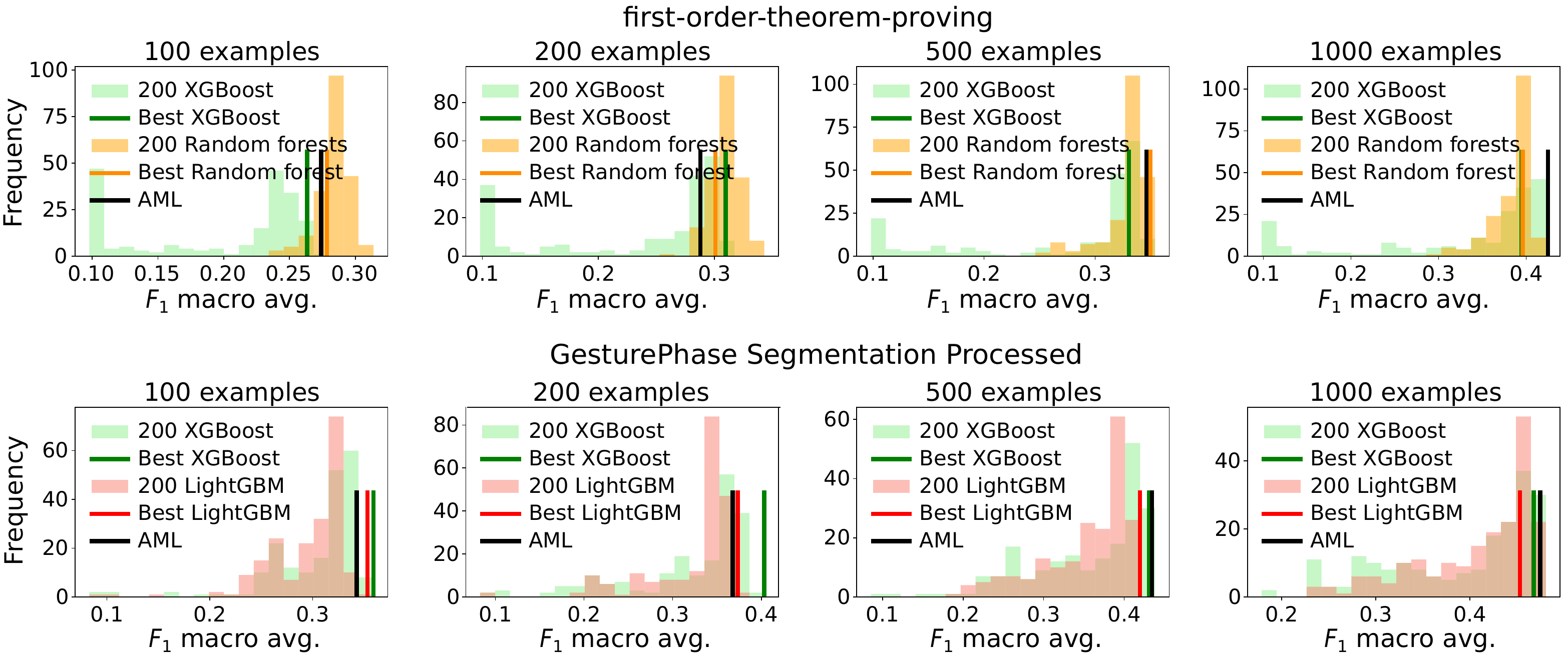}
        \caption{For tabular datasets, examples of distributions of \(F_1\) scores across all runs of the two baseline models that achieved the best results in cross-validation, highlighting the results of the models trained with the hyperparameters selected in cross-validation (labeled best), and the result for AML, for sizes 100, 200, 500, and 1000.}
        \label{fig:tabular_distribution}
    \end{center}
\end{figure}

When looking across the aggregate data, the Friedman test again rejects that the ranks come from the same distribution (\(p<0.001\)), and the post-hoc test (Table~\ref{table:tabular_p_values}) shows AML differs significantly from XGBoost but is comparable (in the sense that the post-hoc tests do not reject equality of distributions) to the remaining methods. Given the negative value of the HL estimator, we have evidence that XGBoost outperforms AML, but in turn we cannot deduce that AML performs differently from other domain-specific methods such as LightGBM and random forests. These relationships are summarized in the critical-difference diagram in Figure \ref{fig:times_best_tabular}, bottom.

\begin{table}
  \caption{Results of post-hoc pairwise Wilcoxon signed-rank tests for AML compared with the baseline methods on tabular datasets. Positive HL estimator values indicate AML achieves higher \(F_1\) than the baselines, and vice versa.}
  \label{table:tabular_p_values}
  \centering
  \small
  \begin{tabular}{llllll}
    \toprule
    Method & Raw \(p\)-value & Adjusted \(p\)-value & HL estimator & 95\% CI & Reject \\
    \midrule
        XGBoost & \(<0.001\) & \(<0.001\) & \(-0.0165\) & \([-0.0248, -0.0091]\)  & Yes \\[1pt]
        Random forest & \(0.0757\) & \(0.3029\) & \(-0.0057\) & \([-0.0141, +0.0005]\) & No \\[1pt]
        LightGBM & \(0.2003\) & \(0.6009\) & \(-0.0049\) & \([-0.0109, +0.0025]\) & No \\[1pt]
        SVM & \(0.0275\) & \(0.1375\) & \(+0.0115\) & \([+0.0013, +0.0220]\) & No \\[1pt]
        MLP & \(0.0083\) & \(0.0579\) & \(+0.0139\) & \([+0.0038, +0.0246]\) & No \\
    \bottomrule
  \end{tabular}
\end{table}

The distributions of training-run results for tabular baselines (Figure \ref{fig:tabular_distribution}) show wider spread than in the image case. In the small-data regime, the cross-validation procedure often does not select the best-performing run; for example, on \emph{first-order-theorem-proving} at 100 and 200 examples, XGBoost and random forests average \(F_1\) macro scores of \(0.2835\) (\(\sigma = 0.0116\)) and \(0.3092\) (\(\sigma = 0.0120\)) respectively, with maxima of \(0.3138\) and \(0.3429\), but the cross-validated runs achieve \(0.2784\) and \(0.3008\).

\subsection{The role of the logistic regression readout}\label{section:logistic_regression_role}

We used a logistic regression readout as a way to classify using AML, thus a question arises regarding the role of the AML model in test classification. On image datasets, AML with the logistic regression readout achieves higher \(F_1\) scores than the fewest misses variant on every (dataset, training-size) pair. A Wilcoxon signed-rank test rejects equivalence (\(p<0.001\)), with HL estimator \(+0.0368\) in favor of the readout (95\% CI \([0.0310, 0.0427]\)). This effect size is comparable in magnitude to AML's HL gap over LightGBM in Table~\ref{table:image_p_values} (\(+0.0334\)). If we were to use fewest misses instead of the logistic regression readout as a method for classifying in AML, AML would achieve a mean rank of \(4.7887\), slightly better than LightGBM's \(4.8099\). 
While the readout's contribution is substantial, the fewest misses variant still achieves a mean rank competitive with one of the baselines.

In the case of tabular data, the logistic regression readout achieves a higher \(F_1\) score in 89 of the 155 (dataset, training-size) pairs. The Wilcoxon signed-rank test again rejects equivalency (\(p=0.0016\)), with an HL estimator of \(+0.0056\) in favor of the logistic regression readout (95\% CI \([0.0020, 0.0095]\)). In terms of mean ranks using fewest misses instead of the logistic regression readout, AML would achieve a mean rank of \(3.7516\), above those of MLP and SVM (\(4.1806\) and \(4.0258\) respectively), but below those of the methods that incorporate modality specific biases.

We conclude that the logistic regression readout substantially improves AML's classification performance over the fewest misses alternative, but that the algebraic model on its own already encodes meaningful task-relevant information, achieving a mean rank competitive with several baselines (better than MLP and SVM on tabular). The choice of the logistic regression readout for our headline results in \S\ref{section:image_classification_results} and \S\ref{section:tabular_classification_results} reflects this performance difference.

\section{Discussion}
\label{sec:analysis}

AML learns by building algebraic models of the training set rather than fitting parameters by optimization. This has two consequences in the low- to medium-data regime. First, AML requires no validation set: every labeled example is used for learning. In small-data settings where each example is precious, this is a meaningful efficiency, and it explains part of AML's relative advantage. Second, AML's inductive bias is fixed by the algebraic structure rather than by hyperparameters tuned on data. Cross-validated baselines must select architectures, regularization, and learning rates from limited data. AML sidesteps the problem entirely. This explanation is consistent with the failure mode we observe in modality-tailored baselines. CNNs encode locality and translation equivariance, which should help on images regardless of training size. Yet CNNs only become competitive with AML at 2000 examples (Figure \ref{fig:times_best_images}, top). 

XGBoost has a small but statistically significant advantage over AML on tabular data, with a median paired difference of approximately \(0.0165\) in macro-\(F_1\) (Table \ref{table:tabular_p_values}). This is not surprising: gradient-boosted trees encode strong, well-validated inductive biases for tabular data (additive recursive partitioning, feature-interaction handling, robustness to missing values) and have benefited from years of optimization for this exact problem class. AML, by contrast, encodes only the order relation between numerical values; everything else (categorical structure, feature interactions) must be discovered from data. That AML remains competitive with random forests and LightGBM despite this asymmetry suggests the algebraic encoding is doing real work.

The same algebraic encoding handles two structurally distinct modalities: dense pixel grids (images) and heterogeneous feature tables (tabular data). This is unusual. AML's success across both modalities, without architectural changes, suggests the algebraic structure is modality-agnostic.
A practical consequence is that a single AML model could in principle integrate features from multiple modalities without architectural redesign. We do not test this here, but the modality-agnostic encoding is a prerequisite for it, and combining heterogeneous inputs into a single algebraic structure is a direction worth exploring.

\section{Limitations}
\label{sec:limitations}

We compare against standard supervised baselines trained from scratch, not few-shot, meta-learning, or transfer-learning methods that leverage auxiliary data or pretrained
representations. We also do not explore the entire possible hyperparameter space for these models, or other potential ways to improve results for parametrized learners such as ensembles of methods.

AML's reported performance uses a logistic regression readout on top of the algebraic atomization; \S\ref{section:logistic_regression_role} reports a comparison with the fewest misses classification method that more directly uses the information from the underlying algebra and shows that the algebraic model encodes meaningful task-relevant information independently of the readout. Nonetheless, our experiments do not fully isolate the readout's contribution from that of the atomization, nor do they disentangle those of the algebraic encoding, rule-finding bias, and the absence of cross-validation; controlled ablations are needed. AML's task encoding is part of its inductive bias, and identifying the most effective encodings for these and new problem classes remains open.

In a similar line, the symbolic algebraic encoding necessary for AML is well-suited to supervised classification tasks, but implementing other kinds of tasks often requires significant work to build tailor-made algebraic embeddings.

Our aggregated statistical analysis treats (dataset, training-size) pairs as approximately independent observations. Performance on the same dataset at adjacent sizes is correlated, which may make our \(p\)-values mildly anti-conservative.

Finally, the current implementation of Sparse Crossing prioritizes a direct realization of the algebraic construction over runtime efficiency. Runtime and memory use for images with dimensionality higher than what we analyze here grow beyond the capabilities of the desktop-class systems available to us. Larger image datasets may require algorithmic acceleration, parallelization, or dedicated hardware implementations. We view this primarily as a systems challenge for a new algebraic learning framework: AML has not yet benefited from the extensive optimization effort that has made deep learning, gradient-boosted trees, and other parametrized methods efficient. Closing this gap is an important direction for future work.

\section{Conclusion}
\label{sec:conclusion}

We have shown that Algebraic Machine Learning~\cite{Maroto,SecondPaperArX,ThirdPaperArX} --- a symbolic framework that learns through algebraic decomposition rather than numerical parameter fitting — is the best-performing method on image classification across the 50–2000 example range we evaluate, and is competitive with domain-specific tabular methods such as random forests and LightGBM in the same range, though XGBoost outperforms it on tabular data overall. Because the same encoding framework is used across both modalities, these results point to a generic algebraic inductive bias that is effective when labeled data are scarce, helped by AML's lack of hyperparameter tuning and validation requirements.

The methods AML competes with have benefited from decades of architectural refinement, optimized implementations, and hardware co-design. AML, by contrast, has a single unoptimized implementation, generic encodings, and no task-specific engineering. We view its current competitiveness as evidence that algebraic decomposition is a viable learning mechanism in the regimes we test; whether and how the framework scales to larger data, additional modalities, or more diverse task types is an empirical question that future work will need to answer.

\section*{Acknowledgments}

We thank Antonio Ricciardo, Nabil Abderrahaman Elena and Emilio Suarez Canedo for discussions. We are grateful for the support from Champalimaud Foundation (Lisbon, Portugal), from FCT -- Fundação para a Ciência e a Tecnologia -- in the context of the project UIDB/04443/2025, and from the European Commission provided through projects H2020 ICT48 \emph{Humane AI; Toward AI Systems That Augment and Empower Humans by Understanding Us, our Society and the World Around Us} (grant \(\# 820437\)) and the H2020 ICT48 project \emph{ALMA: Human Centric Algebraic Machine Learning} (grant \(\# 952091\)).

{
\bibliographystyle{unsrt}
\small 
\bibliography{small_datasets}

@inproceedings{Coates2011AnAO,
  title={An Analysis of Single-Layer Networks in Unsupervised Feature Learning},
  author={Adam Coates and Andrew Ng and Honglak Lee},
  booktitle={International Conference on Artificial Intelligence and Statistics},
  year={2011},
  url={https://api.semanticscholar.org/CorpusID:308212}
}

@book{Burris,
author = {Burris, Stanley and Sankappanavar, H. P.},
isbn = {0387905782},
pages = {276},
publisher = {Springer-Verlag},
title = {{A course in universal algebra}},
year = {1981}
}

@book{Davey_Priestley_2002, place={Cambridge}, edition={2}, title={Introduction to Lattices and Order}, publisher={Cambridge University Press}, author={Davey, B. A. and Priestley, H. A.}, year={2002}}

@article{lecun2010mnist,
  title={MNIST handwritten digit database},
  author={LeCun, Yann and Cortes, Corinna and Burges, CJ},
  journal={ATT Labs [Online]. Available: http://yann.lecun.com/exdb/mnist},
  volume={2},
  year={2010}
}

@article{clanuwat2018deep,
  author       = {Tarin Clanuwat and Mikel Bober-Irizar and Asanobu Kitamoto and Alex Lamb and Kazuaki Yamamoto and David Ha},
  title        = {Deep Learning for Classical Japanese Literature},
  date         = {2018-12-03},
  year         = {2018},
  eprintclass  = {cs.CV},
  eprinttype   = {arXiv},
  eprint       = {cs.CV/1812.01718},
  journal={arXiv:1812.01718},
}

@article{yang2023medmnist,
  title={Medmnist v2--a large-scale lightweight benchmark for 2d and 3d biomedical image classification},
  author={Yang, Jiancheng and Shi, Rui and Wei, Donglai and Liu, Zequan and Zhao, Lin and Ke, Bilian and Pfister, Hanspeter and Ni, Bingbing},
  journal={Scientific Data},
  volume={10},
  number={1},
  pages={41},
  year={2023},
  publisher={Nature Publishing Group UK London}
}

@article{xiao17,
      title={Fashion-MNIST: a Novel Image Dataset for Benchmarking Machine Learning Algorithms}, 
      author={Han Xiao and Kashif Rasul and Roland Vollgraf},
      year={2017},
      eprint={1708.07747},
      archivePrefix={arXiv},
      primaryClass={cs.LG},
      url={https://arxiv.org/abs/1708.07747},
      journal={arXiv:1708.07747}
}

@techreport{Kri09,
            author={Alex Krizhevsky},
            title={Learning multiple layers of features from tiny images},
            year={2009}
}

@misc{aerial-cactus-identification,
    author = {Will Cukierski},
    title = {Aerial Cactus Identification},
    year = {2019},
    howpublished = {\url{https://kaggle.com/competitions/aerial-cactus-identification}},
    note = {Kaggle}
}

@inproceedings{Netzer2011ReadingDI,
  title={Reading Digits in Natural Images with Unsupervised Feature Learning},
  author={Yuval Netzer and Tao Wang and Adam Coates and A. Bissacco and Bo Wu and A. Ng},
  year={2011},
  url={https://api.semanticscholar.org/CorpusID:16852518}
}

@InProceedings{CAVE_0188,
author = {Nene, S.A. and Nayar, S.K. and Murase, H.},
title = {{C}olumbia {O}bject {I}mage {L}ibrary ({C}{O}{I}{L}-20)},
booktitle = {Technical Report, Department of Computer Science, Columbia University CUCS-005-96},
month = {Feb},
year = {1996}
}

@article{Hollmann2025,
  title = {Accurate predictions on small data with a tabular foundation model},
  volume = {637},
  ISSN = {1476-4687},
  url = {http://dx.doi.org/10.1038/s41586-024-08328-6},
  DOI = {10.1038/s41586-024-08328-6},
  number = {8045},
  journal = {Nature},
  publisher = {Springer Science and Business Media LLC},
  author = {Hollmann,  Noah and M\"{u}ller,  Samuel and Purucker,  Lennart and Krishnakumar,  Arjun and K\"{o}rfer,  Max and Hoo,  Shi Bin and Schirrmeister,  Robin Tibor and Hutter,  Frank},
  year = {2025},
  month = jan,
  pages = {319–326}
}

@article{Birkhoff,
 title={Subdirect products in Universal Algebra},
 author={Garrett Birkhoff},
 journal={Bull. Amer. Math. Soc.},
 volume={50},
 pages={764-768},
 year={1944}
}

@article{Maroto,
 title={Algebraic Machine Learning},
 author={Fernando Martin-Maroto and Gonzalo G. de Polavieja},
 journal={arXiv:1803.05252},
 year={2018}
}

@article{SecondPaperArX,
 title={Finite Atomized Semilattices},
 author={Martin-Maroto, Fernando and de Polavieja, Gonzalo G},
 journal={arXiv:2102.08050},
 year={2021}
}

@article{ThirdPaperArX,
 title={Semantic Embeddings in Semilattices},
 author={Martin-Maroto, Fernando and de Polavieja, Gonzalo G},
 journal={arXiv:2205.12618},
 year={2022}
}

@inproceedings{Feigenbaum77,
  title={The Art of Artificial Intelligence: Themes and Case Studies of Knowledge Engineering},
  author={Edward A. Feigenbaum},
  booktitle={International Joint Conference on Artificial Intelligence},
  year={1977},
  url={https://api.semanticscholar.org/CorpusID:1431789}
}

@book{Hayes83,
author = {Hayes-Roth, Frederick and Waterman, Donald A. and Lenat, Douglas B.},
title = {Building expert systems},
year = {1983},
isbn = {0201106868},
publisher = {Addison-Wesley Longman Publishing Co., Inc.},
address = {USA}
}

@article{NewellSimon1976,
author = {Newell, Allen and Simon, Herbert A.},
title = {Computer science as empirical inquiry: symbols and search},
year = {1976},
issue_date = {March 1976},
publisher = {Association for Computing Machinery},
address = {New York, NY, USA},
volume = {19},
number = {3},
issn = {0001-0782},
url = {https://doi.org/10.1145/360018.360022},
doi = {10.1145/360018.360022},
journal = {Commun. ACM},
month = mar,
pages = {113–126},
numpages = {14}
}

@book{bishop2006prml,
author = {Bishop, Christopher M.},
title = {Pattern Recognition and Machine Learning (Information Science and Statistics)},
year = {2006},
isbn = {0387310738},
publisher = {Springer-Verlag},
address = {Berlin, Heidelberg}
}

@book{goodfellow2016deep,
    title={Deep Learning},
    author={Ian Goodfellow and Yoshua Bengio and Aaron Courville},
    publisher={MIT Press},
    note={\url{http://www.deeplearningbook.org}},
    year={2016}
}

@article{lecun1998gradient,
  author={Lecun, Y. and Bottou, L. and Bengio, Y. and Haffner, P.},
  journal={Proceedings of the IEEE}, 
  title={Gradient-based learning applied to document recognition}, 
  year={1998},
  volume={86},
  number={11},
  pages={2278-2324},
  doi={10.1109/5.726791}}

@article{chen2016xgboost,
  title={XGBoost: A Scalable Tree Boosting System},
  author={Tianqi Chen and Carlos Guestrin},
  journal={Proceedings of the 22nd ACM SIGKDD International Conference on Knowledge Discovery and Data Mining},
  year={2016},
  url={https://api.semanticscholar.org/CorpusID:4650265}
}

@inproceedings{ke2017lightgbm,
author = {Ke, Guolin and Meng, Qi and Finley, Thomas and Wang, Taifeng and Chen, Wei and Ma, Weidong and Ye, Qiwei and Liu, Tie-Yan},
title = {LightGBM: a highly efficient gradient boosting decision tree},
year = {2017},
isbn = {9781510860964},
publisher = {Curran Associates Inc.},
address = {Red Hook, NY, USA},
booktitle = {Proceedings of the 31st International Conference on Neural Information Processing Systems},
pages = {3149–3157},
numpages = {9},
location = {Long Beach, California, USA},
series = {NIPS'17}
}

@article{Breiman2001RandomForests,
  title     = "Random forests",
  author    = "Breiman, Leo",
  journal   = "Mach. Learn.",
  publisher = "Springer Science and Business Media LLC",
  volume    =  45,
  number    =  1,
  pages     = "5--32",
  month     =  oct,
  year      =  2001,
  copyright = "https://www.springernature.com/gp/researchers/text-and-data-mining",
  language  = "en"
}

@book{Hollander2015,
  title = {Nonparametric Statistical Methods},
  ISBN = {9781119196037},
  ISSN = {1940-6347},
  url = {http://dx.doi.org/10.1002/9781119196037},
  DOI = {10.1002/9781119196037},
  journal = {Wiley Series in Probability and Statistics},
  publisher = {Wiley},
  author = {Hollander,  Myles and A. Wolfe,  Douglas and Chicken,  Eric},
  year = {2015},
}

@article{cortes1995svm,
  author = {Cortes, Corinna and Vapnik, Vladimir},
  doi = {10.1007/BF00994018},
  journal = {Machine learning},
  number = 3,
  pages = {273--297},
  title = {Support-vector networks},
  volume = 20,
  year = 1995
}

@inproceedings{finn2017model,
author = {Finn, Chelsea and Abbeel, Pieter and Levine, Sergey},
title = {Model-agnostic meta-learning for fast adaptation of deep networks},
year = {2017},
publisher = {JMLR.org},
booktitle = {Proceedings of the 34th International Conference on Machine Learning - Volume 70},
pages = {1126–1135},
numpages = {10},
location = {Sydney, NSW, Australia},
series = {ICML'17}
}

@inproceedings{chen2020simple,
author = {Chen, Ting and Kornblith, Simon and Norouzi, Mohammad and Hinton, Geoffrey},
title = {A simple framework for contrastive learning of visual representations},
year = {2020},
publisher = {JMLR.org},
booktitle = {Proceedings of the 37th International Conference on Machine Learning},
articleno = {149},
numpages = {11},
series = {ICML'20}
}

@article{RiegelEtAl2020,
  author       = {Ryan Riegel and
                  Alexander G. Gray and
                  Francois P. S. Luus and
                  Naweed Khan and
                  Ndivhuwo Makondo and
                  Ismail Yunus Akhalwaya and
                  Haifeng Qian and
                  Ronald Fagin and
                  Francisco Barahona and
                  Udit Sharma and
                  Shajith Ikbal and
                  Hima Karanam and
                  Sumit Neelam and
                  Ankita Likhyani and
                  Santosh K. Srivastava},
  title        = {Logical Neural Networks},
  journal      = {arXiv:2006.13155},
  year         = {2020},
  url          = {https://arxiv.org/abs/2006.13155},
  eprinttype   = {arXiv},
  primaryClass = {cs.AI},
  eprint       = {2006.13155},
}

@article{BadreddineEtAl2022,
title = {Logic Tensor Networks},
journal = {Artificial Intelligence},
volume = {303},
pages = {103649},
year = {2022},
issn = {0004-3702},
doi = {https://doi.org/10.1016/j.artint.2021.103649},
url = {https://www.sciencedirect.com/science/article/pii/S0004370221002009},
author = {Samy Badreddine and Artur {d'Avila Garcez} and Luciano Serafini and Michael Spranger}
}

@inproceedings{CohenWelling2016,
author = {Cohen, Taco S. and Welling, Max},
title = {Group equivariant convolutional networks},
year = {2016},
publisher = {JMLR.org},
booktitle = {Proceedings of the 33rd International Conference on International Conference on Machine Learning - Volume 48},
pages = {2990–2999},
numpages = {10},
location = {New York, NY, USA},
series = {ICML'16}
}

@book{DrtonSturmfelsSullivant2009,
  title = {Lectures on Algebraic Statistics},
  ISBN = {9783764389055},
  url = {http://dx.doi.org/10.1007/978-3-7643-8905-5},
  DOI = {10.1007/978-3-7643-8905-5},
  publisher = {Birkh\"{a}user Basel},
  author = {Drton,  Mathias and Sturmfels,  Bernd and Sullivant,  Seth},
  year = {2009}
}

@book{FongSpivak2018, place={Cambridge}, title={An Invitation to Applied Category Theory: Seven Sketches in Compositionality}, publisher={Cambridge University Press}, author={Fong, Brendan and Spivak, David I.}, year={2019}}

@article{Demsar06,
author = {Dem\v{s}ar, Janez},
title = {Statistical Comparisons of Classifiers over Multiple Data Sets},
year = {2006},
issue_date = {12/1/2006},
publisher = {JMLR.org},
volume = {7},
issn = {1532-4435},
journal = {J. Mach. Learn. Res.},
month = dec,
pages = {1–30},
numpages = {30}
}

@article{GarciaetalExtensionDemsar,
  author  = {Salvador Garc{{\'i}}a and Francisco Herrera},
  title   = {An Extension on ``Statistical Comparisons of Classifiers over Multiple Data Sets'' for all Pairwise Comparisons},
  journal = {Journal of Machine Learning Research},
  year    = {2008},
  volume  = {9},
  number  = {89},
  pages   = {2677--2694},
  url     = {http://jmlr.org/papers/v9/garcia08a.html}
}

@article{arxiv2025,
      title={Algebraic Machine Learning: Learning as computing an algebraic decomposition of a task}, 
      author={Fernando Martin-Maroto and Nabil Abderrahaman and David Mendez and Gonzalo G. de Polavieja},
      year={2025},
      journal={arXiv:2502.19944},
      eprint={2502.19944},
      archivePrefix={arXiv},
      primaryClass={cs.LG},
      url={https://arxiv.org/abs/2502.19944}, 
}

@article{Benavoli16,
  author  = {Alessio Benavoli and Giorgio Corani and Francesca Mangili},
  title   = {Should We Really Use Post-Hoc Tests Based on Mean-Ranks?},
  journal = {Journal of Machine Learning Research},
  year    = {2016},
  volume  = {17},
  number  = {5},
  pages   = {1--10},
  url     = {http://jmlr.org/papers/v17/benavoli16a.html}
}

@article{Holm79,
 ISSN = {03036898, 14679469},
 URL = {http://www.jstor.org/stable/4615733},
 author = {Sture Holm},
 journal = {Scandinavian Journal of Statistics},
 number = {2},
 pages = {65--70},
 publisher = {[Board of the Foundation of the Scandinavian Journal of Statistics, Wiley]},
 title = {A Simple Sequentially Rejective Multiple Test Procedure},
 urldate = {2026-04-23},
 volume = {6},
 year = {1979}
}

@article{bloodmnist,
title = {A dataset of microscopic peripheral blood cell images for development of automatic recognition systems},
journal = {Data in Brief},
volume = {30},
pages = {105474},
year = {2020},
issn = {2352-3409},
doi = {https://doi.org/10.1016/j.dib.2020.105474},
url = {https://www.sciencedirect.com/science/article/pii/S2352340920303681},
author = {Andrea Acevedo and Anna Merino and Santiago Alférez and Ángel Molina and Laura Boldú and José Rodellar},
}

@article{organcmnist,
  author={Xu, Xuanang and Zhou, Fugen and Liu, Bo and Fu, Dongshan and Bai, Xiangzhi},
  journal={IEEE Transactions on Medical Imaging}, 
  title={Efficient Multiple Organ Localization in CT Image Using 3D Region Proposal Network}, 
  year={2019},
  volume={38},
  number={8},
  pages={1885-1898},
  doi={10.1109/TMI.2019.2894854}}

@article{pneumoniamnist,
title = {Identifying Medical Diagnoses and Treatable Diseases by Image-Based Deep Learning},
journal = {Cell},
volume = {172},
number = {5},
pages = {1122-1131.e9},
year = {2018},
issn = {0092-8674},
doi = {https://doi.org/10.1016/j.cell.2018.02.010},
url = {https://www.sciencedirect.com/science/article/pii/S0092867418301545},
author = {Daniel S. Kermany and Michael Goldbaum and Wenjia Cai and Carolina C.S. Valentim and Huiying Liang and Sally L. Baxter and Alex McKeown and Ge Yang and Xiaokang Wu and Fangbing Yan and Justin Dong and Made K. Prasadha and Jacqueline Pei and Magdalene Y.L. Ting and Jie Zhu and Christina Li and Sierra Hewett and Jason Dong and Ian Ziyar and Alexander Shi and Runze Zhang and Lianghong Zheng and Rui Hou and William Shi and Xin Fu and Yaou Duan and Viet A.N. Huu and Cindy Wen and Edward D. Zhang and Charlotte L. Zhang and Oulan Li and Xiaobo Wang and Michael A. Singer and Xiaodong Sun and Jie Xu and Ali Tafreshi and M. Anthony Lewis and Huimin Xia and Kang Zhang},
}

@article{dermamnist1,
  title = {The HAM10000 dataset,  a large collection of multi-source dermatoscopic images of common pigmented skin lesions},
  volume = {5},
  ISSN = {2052-4463},
  url = {http://dx.doi.org/10.1038/sdata.2018.161},
  DOI = {10.1038/sdata.2018.161},
  number = {1},
  journal = {Scientific Data},
  publisher = {Springer Science and Business Media LLC},
  author = {Tschandl,  Philipp and Rosendahl,  Cliff and Kittler,  Harald},
  year = {2018},
  month = Aug 
}

@article{dermamnist2,
      title={Skin Lesion Analysis Toward Melanoma Detection 2018: A Challenge Hosted by the International Skin Imaging Collaboration (ISIC)}, 
      author={Noel Codella and Veronica Rotemberg and Philipp Tschandl and M. Emre Celebi and Stephen Dusza and David Gutman and Brian Helba and Aadi Kalloo and Konstantinos Liopyris and Michael Marchetti and Harald Kittler and Allan Halpern},
      year={2019},
      eprint={1902.03368},
      archivePrefix={arXiv},
      primaryClass={cs.CV},
      url={https://arxiv.org/abs/1902.03368}, 
      journal={arXiv:1902.03368}
}

@article{CactusAerial,
title = {Columnar cactus recognition in aerial images using a deep learning approach},
journal = {Ecological Informatics},
volume = {52},
pages = {131-138},
year = {2019},
issn = {1574-9541},
doi = {https://doi.org/10.1016/j.ecoinf.2019.05.005},
url = {https://www.sciencedirect.com/science/article/pii/S1574954119300895},
author = {Efren López-Jiménez and Juan Irving Vasquez-Gomez and Miguel Angel Sanchez-Acevedo and Juan Carlos Herrera-Lozada and Abril Valeria Uriarte-Arcia}
}

@Inbook{automl2018,
author="Guyon, Isabelle
and Sun-Hosoya, Lisheng
and Boull{\'e}, Marc
and Escalante, Hugo Jair
and Escalera, Sergio
and Liu, Zhengying
and Jajetic, Damir
and Ray, Bisakha
and Saeed, Mehreen
and Sebag, Mich{\`e}le
and Statnikov, Alexander
and Tu, Wei-Wei
and Viegas, Evelyne",
editor="Hutter, Frank
and Kotthoff, Lars
and Vanschoren, Joaquin",
title="Analysis of the AutoML Challenge Series 2015--2018",
bookTitle="Automated Machine Learning: Methods, Systems, Challenges",
year="2019",
publisher="Springer International Publishing",
address="Cham",
pages="177--219",
isbn="978-3-030-05318-5",
doi="10.1007/978-3-030-05318-5_10",
url="https://doi.org/10.1007/978-3-030-05318-5_10"
}

@misc{Australian,
  author       = {Quinlan, Ross},
  title        = {{Statlog (Australian Credit Approval)}},
  year         = {1987},
  howpublished = {UCI Machine Learning Repository},
  note         = {{DOI}: https://doi.org/10.24432/C59012}
}

@article{blood-transfusion,
title = {Knowledge discovery on RFM model using Bernoulli sequence},
journal = {Expert Systems with Applications},
volume = {36},
number = {3, Part 2},
pages = {5866-5871},
year = {2009},
issn = {0957-4174},
doi = {https://doi.org/10.1016/j.eswa.2008.07.018},
url = {https://www.sciencedirect.com/science/article/pii/S0957417408004508},
author = {I-Cheng Yeh and King-Jang Yang and Tao-Ming Ting},
}

@inproceedings{car,
  title={Knowledge acquisition and explanation for multi-attribute decision making},
  author={Bohanec, Marko and Rajkovic, Vladislav},
  booktitle={8th intl workshop on expert systems and their applications},
  pages={59--78},
  year={1988},
  organization={Avignon France}
}

@misc{churn,
author = {Unknown},
title = {Churn},
howpublished = {OpenML Dataset Repository},
note = {{OpenML ID}: 40701}
}

@misc{cmc,
  author       = {Lim, Tjen-Sien},
  title        = {{Contraceptive Method Choice}},
  year         = {1999},
  howpublished = {UCI Machine Learning Repository},
  note         = {{DOI}: https://doi.org/10.24432/C59W2D}
}

@misc{credit-g,
  author       = {Hofmann, Hans},
  title        = {{Statlog (German Credit Data)}},
  year         = {1994},
  howpublished = {UCI Machine Learning Repository},
  note         = {{DOI}: https://doi.org/10.24432/C5NC77}
}

@misc{dna,
  title        = {{Molecular Biology (Splice-junction Gene Sequences)}},
  year         = {1991},
  howpublished = {UCI Machine Learning Repository},
  note         = {{DOI}: https://doi.org/10.24432/C5M888}
}

@article{eucalyptus,
  title={Eucalyptus species selection for soil conservation in seasonally dry hill country - twelfth year assessment.},
  author={Blake Bulloch},
  journal={New Zealand journal of forestry science},
  year={1991},
  volume={21},
  pages={10-31},
  url={https://api.semanticscholar.org/CorpusID:83106110}
}

@misc{first-order_theorem_proving,
  author       = {Bridge, James and Holden, Sean and Paulson, Lawrence},
  title        = {{First-order theorem proving}},
  year         = {2012},
  howpublished = {UCI Machine Learning Repository},
  note         = {{DOI}: https://doi.org/10.24432/C5RC9X}
}

@inproceedings{GesturePhaseSegmentation,
author = {Madeo, Renata C. B. and Lima, Clodoaldo A. M. and Peres, Sarajane M.},
title = {Gesture unit segmentation using support vector machines: segmenting gestures from rest positions},
year = {2013},
isbn = {9781450316569},
publisher = {Association for Computing Machinery},
address = {New York, NY, USA},
url = {https://doi.org/10.1145/2480362.2480373},
doi = {10.1145/2480362.2480373},
booktitle = {Proceedings of the 28th Annual ACM Symposium on Applied Computing},
pages = {46–52},
numpages = {7},
location = {Coimbra, Portugal},
series = {SAC '13}
}

@inproceedings{AutoML15,
title = "Design of the 2015 ChaLearn AutoML challenge",
author = "Isabelle Guyon and Kristin Bennett and Gavin Cawley and Escalante, \{Hugo Jair\} and Sergio Escalera and Ho, \{Tin Kam\} and N{\'u}ria Maci{\`a} and Bisakha Ray and Mehreen Saeed and Alexander Statnikov and Evelyne Viegas",
note = "Publisher Copyright: {\textcopyright} 2015 IEEE.; International Joint Conference on Neural Networks, IJCNN 2015 ; Conference date: 12-07-2015 Through 17-07-2015",
year = "2015",
month = sep,
day = "28",
doi = "10.1109/IJCNN.2015.7280767",
language = "English",
series = "Proceedings of the International Joint Conference on Neural Networks",
booktitle = "2015 International Joint Conference on Neural Networks, IJCNN 2015",
publisher = "IEEE Signal Processing Society",
address = "United States",
}

@INPROCEEDINGS{kc1, author={Nan Niu and Mahmoud, A.}, booktitle={Requirements Engineering Conference (RE), 2012 20th IEEE International}, title={Enhancing candidate link generation for requirements tracing: The cluster hypothesis revisited}, year={2012}, pages={81-90}, doi={10.1109/RE.2012.6345842}, ISSN={1090-750X}, month={Sept},}

@misc{kr-vs-kp,
  author       = {Shapiro, Alen},
  title        = {{Chess (King-Rook vs. King-Pawn)}},
  year         = {1983},
  howpublished = {UCI Machine Learning Repository},
  note         = {{DOI}: https://doi.org/10.24432/C5DK5C}
}

@misc{mfeat-factors,
  author       = {Duin, Robert},
  title        = {{Multiple Features}},
  year         = {1998},
  howpublished = {UCI Machine Learning Repository},
  note         = {{DOI}: https://doi.org/10.24432/C5HC70}
}

@misc{ozone_level_8hr,
  author       = {Zhang, Kun and Fan, Wei and Yuan, XiaoJing},
  title        = {{Ozone Level Detection}},
  year         = {2008},
  howpublished = {UCI Machine Learning Repository},
  note         = {{DOI}: https://doi.org/10.24432/C5NG6W}
}

@INPROCEEDINGS{pc4, author={Menzies, T. and Di Stefano, J.S.}, booktitle={High Assurance Systems Engineering, 2004. Proceedings. Eighth IEEE International Symposium on}, title={How good is your blind spot sampling policy}, year={2004}, pages={129-138}, doi={10.1109/HASE.2004.1281737}, ISSN={1530-2059}, month={March},}

@inproceedings{phoneme,
  title={Esprit II Project 5516 ROARS Robust Analytic Speech Recognition System},
  author={Pierre Alinat and Jean-Marie Pierrel},
  year={1994},
  url={https://api.semanticscholar.org/CorpusID:114700444}
}

@misc{qsar_biodeg,
  author       = {Mansouri, Kamel and Ringsted, Tine and Ballabio, Davide and Todeschini, Roberto and Consonni, Viviana},
  title        = {{QSAR biodegradation}},
  year         = {2013},
  howpublished = {UCI Machine Learning Repository},
  note         = {{DOI}: https://doi.org/10.24432/C5H60M}
}

@misc{Satellite,
author = {Goldstein, Markus},
publisher = {Harvard Dataverse},
title = {{Unsupervised Anomaly Detection Benchmark}},
UNF = {UNF:6:EnytiA6wCIilzHetzQQV7A==},
year = {2015},
version = {V1},
doi = {10.7910/DVN/OPQMVF},
url = {https://doi.org/10.7910/DVN/OPQMVF}
}

@misc{segment,
  title        = {{Image Segmentation}},
  year         = {1990},
  howpublished = {UCI Machine Learning Repository},
  note         = {{DOI}: https://doi.org/10.24432/C5GP4N}
}

@misc{steel_plates_fault,
  author       = {Buscema, M and Terzi, S and Tastle, W},
  title        = {{Steel Plates Faults}},
  year         = {2010},
  howpublished = {UCI Machine Learning Repository},
  note         = {{DOI}: https://doi.org/10.24432/C5J88N}
}

@inproceedings{vehicle,
  title={Vehicle Recognition Using Rule Based Methods, Project Report},
  booktitle={Turing Institute, Glasgow},
  author={Jan Paul Siebert},
  year={1987},
  url={https://api.semanticscholar.org/CorpusID:53782195}
}

@misc{wilt,
  author       = {Johnson, Brian},
  title        = {{Wilt}},
  year         = {2013},
  howpublished = {UCI Machine Learning Repository},
  note         = {{DOI}: https://doi.org/10.24432/C5KS4M}
}

@article{wine-quality-white,
  title={Modeling wine preferences by data mining from physicochemical properties},
  author={P. Cortez and Antonio Lu{\'i}z Cerdeira and Fernando Almeida and Telmo Matos and Jos{\'e} Reis},
  journal={Decis. Support Syst.},
  year={2009},
  volume={47},
  pages={547-553},
  url={https://api.semanticscholar.org/CorpusID:2996254}
}

@misc{yeast,
  author       = {Nakai, Kenta},
  title        = {{Yeast}},
  year         = {1991},
  howpublished = {UCI Machine Learning Repository},
  note         = {{DOI}: https://doi.org/10.24432/C5KG68}
}
}

\newpage
\appendix
\section{Datasets characteristics}\label{section:datasets_chars}

Table \ref{table:image_datasets} contains the characteristics of the analyzed image datasets. The train/validation/test splits are irrelevant for our purposes as we only take examples from the train dataset. All datasets are provided under a license allowing for research use, including CC BY-SA 3.0 (MNIST), CC BY-SA 4.0 (Kuzushiji-49), CC-BY 4.0 (BloodMNIST, OrganCMNIST, PneumoniaMNIST), CC BY-NC 4.0 (DermaMNIST), MIT (Fashion-MNIST, CIFAR-10), CC0 (Street view house number), GPL-2 (Aerial Cactus Identification) and other/unknown licenses allowing for academic use (STL-10, COIL-20).

\begin{table}[h]
  \caption{Characteristics of analyzed image datasets, including their domain, resolution, color channels, the number of test examples, the number of targets and the ratio between the most and least common examples in the test split.}
  \label{table:image_datasets}
  \centering
  \small
  \begin{tabular}{llllll}
    \toprule
    Dataset & Domain & Res.~\& channels\ & Test set & Targets\ & Ratio \\
    \midrule\\[-8pt]
    MNIST~\cite{lecun2010mnist} & Handwriting recognition & \(28\times 28\), 1 & 10000 & \(10\) & \(1.27\) \\[2pt]
    Kuzushiji-49~\cite{clanuwat2018deep} & Handwriting recognition & \(28\times 28\), 1 & 38547 & \(49\) & \(15.63\) \\[2pt]
    BloodMNIST~\cite{bloodmnist, yang2023medmnist} & Medical images & \(28\times 28\), 3 & 3421 & \(8\) & \(2.74\) \\[2pt]
    OrganCMNIST~\cite{organcmnist, yang2023medmnist} & Medical images & \(28\times 28\), 1 & 8216 & \(11\) & 4.36 \\[2pt]
    PneumoniaMNIST~\cite{pneumoniamnist, yang2023medmnist} & Medical images & \(28\times 28\), 1 & 624 & \(2\) & \(1.67\) \\[2pt]
    DermaMNIST~\cite{dermamnist1, dermamnist2, yang2023medmnist} & Medical images & \(28 \times 28\), 3 & 2005 & \(8\) & \(2.74\) \\[2pt]
    CIFAR-10~\cite{Kri09} & Object recognition & \(32\times 32\), 3 & 10000 & \(10\) & \(1.00\) \\[2pt]
    Fashion-MNIST~\cite{xiao17} & Object recognition & \(28\times 28\), 1 & 10000 & \(10\) & \(1.00\) \\[2pt]
    STL-10~\cite{Coates2011AnAO} & Object recognition & \(32\times 32\), 3 & 8000 & \(10\) & \(1.00\) \\[2pt]
    Aerial cactus & Object recognition & \(32\times 32\), 3 & 10000 & \(2\) & \(3.09\) \\
    identification~\cite{aerial-cactus-identification, CactusAerial} & & & & & \\[2pt]
    Street view & Number recognition & \(32\times 32\), 3 & 26032 & \(10\) & \(3.20\) \\
    house number~\cite{Netzer2011ReadingDI} & & & & & \\[2pt]
    COIL-20~\cite{CAVE_0188} & Object recognition & \(128\times128\), 1 & 440 & 20 & \(1.00\)  \\[1pt]
    \bottomrule
  \end{tabular}
\end{table}

Table \ref{table:tabular_datasets} contains the characteristics of the analyzed tabular datasets. All datasets were acquired from OpenML and are displayed using their names in OpenML; OpenML identifiers are available in \cite[Extended Data Table 3]{Hollmann2025}. Many of these datasets are part of the UCI Machine Learning Repository and are available under the CC-BY 4.0 license (Australian, blood-transfusion, car, cmc, credit-g, dna, first-order-theorem-proving, Gesture Phase Segmentation Processed, kr-vs-kp, mfeat-factors, ozone-level-8hr, qsar-biodeg, segment, steel-plates-fault, vehicle, wilt, yeast). Several datasets have origin in the AutoML challenges and do not have an explicit license, but are nonetheless of frequent use in research (ada, jasmine, madeline, philippine, sylvine). Satellite is released under the CC0 license, and wine-quality-white under the Database Contents License (DbCL) 1.0. The datasets kc1, pc4 and phoneme originate from old research projects (NASA PROMISE for kc1 and pc4; EU ELENA for phoneme) and have no disclosed licenses. Finally, churn and eucalyptus have no known license. All datasets are of frequent use in research, but those with undisclosed licenses may not be suitable for commercial use.

\begin{table}[h]
  \caption{Characteristics of analyzed tabular datasets, including their domain, number of features, how many of them are categorical, the number of test examples, the number of targets and the ratio between the most and least common examples in the test split.}
  \label{table:tabular_datasets}
  \centering
  \small
  \begin{tabular}{lllllll}
    \toprule
    Dataset     & Domain & Feats.\ & Cat.\ feats.\ & Test set & Targets\ & Ratio \\
    \midrule
    ada \cite{automl2018} & Census & 48 & 0 & 415 & 2 & 2.71\\[2pt]
    Australian \cite{Australian} & Finance & 14 & 8 & 69 & 2 & 1.03 \\[2pt]
    blood-transfusion & Healthcare & 4 & 0 & 75 & 2 & 3.41 \\
    -service-center \cite{blood-transfusion} & & & & & \\[2pt]
    car \cite{car} & Automotive & 6 & 6 & 173 & 4 & 20.67 \\[2pt]
    churn \cite{churn} & Telecommunication & 20 & 4 & 500 & 2 & 5.41 \\[2pt]
    cmc \cite{cmc} & Public health & 9 & 7 & 148 & 3 & 2.00 \\[2pt]
    credit-g \cite{credit-g} & Finance & 20 & 13 & 100 & 2 & 1.63 \\[2pt]
    dna \cite{dna} & Biology & 180 & 180 & 319 & 3 & 2.41 \\[2pt]
    eucalyptus \cite{eucalyptus} & Agriculture & 19 & 5 & 74 & 5 & 1.5 \\[2pt]
    first-order & Computational & 51 & 0 & 612 & 6 & 5.1 \\
    -theorem-proving \cite{first-order_theorem_proving} & logic & & & & & \\[2pt]
    Gesture Phase segmen- & Human-computer & 32 & 0 & 988 & 5 & 3.82 \\
    tation processed \cite{GesturePhaseSegmentation} & interaction & & & & & \\[2pt]
    jasmine \cite{AutoML15} & Natural language & 144 & 136 & 299 & 2 & 1.03 \\
    & processing & & & & & \\[2pt]
    kc1 \cite{kc1} & Software engineering & 21 & 0 & 211 & 2 & 4.86 \\[2pt]
    kr-vs-kp \cite{kr-vs-kp} & Game strategy & 36 & 36 & 320 & 2 & 1.03 \\[2pt]
    madeline \cite{AutoML15} & Synthetic & 259 & 0 & 314 & 2 & 1.11 \\[2pt]
    mfeat-factors \cite{mfeat-factors} & Handwriting recognition & 216 & 0 & 200 & 10 & 1.86 \\[2pt]
    ozone-level-8hr \cite{ozone_level_8hr} & Environmental & 72 & 0 & 254 & 2 & 20.17 \\[2pt]
    pc4 \cite{pc4} & Software engineering & 37 & 0 & 146 & 2 & 9.43 \\[2pt]
    philippine \cite{AutoML15} & Bioinformatics & 308 & 0 & 584 & 2 & 1.01 \\[2pt]
    phoneme \cite{phoneme} & Audio & 5 & 0 & 541 & 2 & 2.66 \\[2pt]
    qsar-biodeg \cite{qsar_biodeg} & Environmental & 41 & 0 & 106 & 2 & 2.12 \\[2pt]
    Satellite \cite{Satellite} & Environmental Science & 36 & 0 & 510 & 2 & 126.5 \\[2pt]
    segment \cite{segment} & Computer Vision & 16 & 0 & 231 & 7 & 1.37 \\[2pt]
    steel-plates-fault \cite{steel_plates_fault} & Industrial & 27 & 0 & 195 & 7 & 17.5 \\[2pt]
    sylvine \cite{AutoML15} & Environmental Science & 20 & 0 & 513 & 2 & 1.06 \\[2pt]
    vehicle \cite{vehicle} & Image classification & 18 & 0 & 85 & 4 & 1.63 \\[2pt]
    wilt \cite{wilt} & Environmental & 5 & 0 & 484 & 2 & 19.17 \\[2pt]
    wine-quality-white \cite{wine-quality-white} & Food and beverage & 11 & 0 & 490 & 7 & 105.00 \\[2pt]
    yeast \cite{yeast} & Biology & 8 & 0 & 149 & 10 & 25.00 \\[1pt]
    \bottomrule
  \end{tabular}
\end{table}

\section{Methodology for experiments}\label{section:experiment_methodology}

\subsection{Baseline methods}\label{section:baselines_methods}

In both image and tabular datasets, hyperparameters for the final parametrized models were chosen via stratified 5-fold cross-validation, using Scikit-learn, with 200 runs per method. Hyperparameters were sampled randomly within reasonable ranges. Image datasets were normalized so that the train dataset in each case has a mean of 0 and a standard deviation of 1. For tabular datasets, and for methods that do not admit categorical features (MLPs, SVMs), each possible value of the categorical features was mapped to a fixed integer value. For these methods, the data was then normalized so that each variable in the training dataset has a mean of 0 and a standard deviation of 1. We now outline, for each of the methods, the computational tools and parameter ranges sampled; any omitted hyperparameter was left as default in the corresponding software libraries.

\paragraph{XGBoost.} Computations were performed using the Python \texttt{xgboost} library's Scikit-learn API. All parameters were uniformly sampled, other than the learning rate, which was sampled from a loguniform distribution. The parameter ranges are \texttt{colsample\_bytree} from \(0.6\) to \(1\), \texttt{max\_depth} from \(3\) to \(20\), gamma from \(0\) to \(0.01\), learning rate/\texttt{eta} from \(0.001\) to \(0.2\), \texttt{n\_estimators} from \(50\) to \(500\), \texttt{reg\_alpha} from \(0\) to \(2\), \texttt{reg\_lambda} from \(0\) to \(2\) and subsample from \(0.6\) to \(1\).

\paragraph{LightGBM.} Computations were similarly performed using the Python \texttt{lightgbm} library's Scikit-learn API. Considered ranges are the same as those used in XGBoost, and \texttt{num\_leaves} was sampled in the range from 20 to 50.

\paragraph{Random forests.} We used Scikit-learn's \texttt{RandomForestClassifier}. All parameters are uniformly sampled, with ranges \texttt{max\_depth} from \(3\) to \(20\), \texttt{n\_estimators} from \(50\) to \(500\), \texttt{min\_samples\_split} from \(2\) to \(10\), and \texttt{min\_samples\_leaf} from \(1\) to \(4\).

\paragraph{SVMs.} Computations were performed using the \texttt{SVC} method from Scikit-learn. The kernel is randomly sampled for each run between linear, radial basis function (rbf), sigmoid and polynomial. The regularization parameter \(C\) is sampled to be loguniform in the range of \(0.1\) to \(100\). For sigmoid, rbf and polynomial kernels, \texttt{gamma} is sampled from a loguniform in the range \(0.001\) to \(10\). For the polynomial kernel, the degree is uniformly sampled in the range from \(2\) to \(5\). We used \texttt{l2} as the penalty.

\paragraph{MLPs.} MLP computations were made using the \texttt{skorch} library, providing an API between Pytorch and Scikit-learn, thus allowing for the usage of Scikit-learn's cross-validation methods. Learning rate was uniformly sampled in the interval \((10^{-5}, 10^{-4})\), and layer sizes were uniformly selected from 128, 256, 512, 2048 or 4096 for the first hidden layer, 128, 256, 1024 or 2048 for the second hidden layer, 0, 128, 256 or 512 for the third hidden layer and 0, 128 or 256 for the fourth hidden layer. Thus, on average we can expect 50 models to be two layers deep, 100 models to be three layers deep and 50 models to be four layers deep. Intermediate layers use ReLU as the activation function. Training was done for a maximum of 50 epochs, and the epoch with the best validation accuracy was recorded for each run. The final model for which metrics are reported is trained with the hyperparameters that achieved the best cross-validation results, for a number of epochs equal to the mean of the epochs used in cross-validation. The models were trained using the ADAM optimizer with a batch size of 128, with the parameters \texttt{betas} and \texttt{eps} left as default in Pytorch.

\paragraph{CNNs.} CNN computations were performed only on image datasets, with the same tooling used for MLPs, and we used the same learning rate interval, optimizer, and batch size. We used a linear classifier of fixed size with 2 hidden layers, of sizes 128 and 64. We used two to three convolutional layers, with square kernel sizes sampled from 3, 5 and 7 for the first two layers, and from 0, 3 and 5 for the third layer. We can thus expect an average of 100 runs with 2 convolutional layers, and another 100 runs with 3 layers. ReLU was used for the activation functions, and intermediate convolutional layers use dropout with a value of \(0.2\) for normalization.

Other than the SVMs, all methods were trained minimizing cross-entropy loss. The reported best test results correspond to a model trained on the entire train dataset, using the hyperparameters of the model that achieved the best average results between the five cross-validation runs. To ensure train samples were balanced between the different classes, for all methods except for MLPs and CNNs, train samples used at each epoch were sampled from the corresponding train dataset with a weight corresponding to their inverse prevalence in the dataset, whereas for MLPs and CNNs the loss was scaled according to the inverse prevalence of each of the classes.

All baseline computations were performed on a system with an AMD Epyc 7301 CPU with 128 GB of RAM and an NVIDIA GTX 1080 Ti GPU. GPU acceleration was used for XGBoost, MLPs and CNNs, the remaining methods were run on CPU with a number of workers equal to the number of CPU physical threads. RAM usage was not observed to surpass 32 GB during the course of the experiments. Run times varied substantially between datasets and training set sizes, ranging from a few seconds to around 24 hours per dataset and training set size, mostly when running XGBoost on datasets with 2000 train examples.

\subsection{AML}\label{section:aml_methods}

The general procedure and parameter selection (batch number, batch size growth, post-training model reduction, logistic regression readout) is introduced in \S\ref{section:aml_computations}, and except for batch sizes mostly matches \cite[Methods]{arxiv2025}. For our experiments we used an internal development version of Algebraic AI's Open AML Engine (\href{https://github.com/Algebraic-AI/Open-AML-Engine}{https://github.com/Algebraic-AI/Open-AML-Engine}) with additional optimizations that reduce runtime particularly when dealing with numerical values and their order relations. Any hyperparameters mentioned in \cite[Methods]{arxiv2025} not disclosed in \S\ref{section:aml_computations} (simplification threshold, union model fractioning parameter) were left as their default for image classification tasks (\(1.5\) and \(0.1\) respectively).

All experiments were run on a system with an AMD Ryzen 5 5950X CPU with 128 GB of RAM and an NVIDIA RTX 3090 GPU. The Open AML engine does not make use of GPU acceleration and is single threaded in a lot of its tasks, so we typically ran two datasets concurrently. The GPU was used to train the logistic regression readout, which was trained using Pytorch with the ADAM optimizer.

Overall runtime per dataset varied widely between different datasets and training set sizes, ranging from a few minutes to around one week per dataset and training set size, mainly when running image datasets with 2000 train examples. Tabular datasets often run faster, typically finishing an entire train and evaluation run with 2000 train examples in 48 to 72 hours. We did not observe RAM usage surpassing 64 GB for a single dataset. Most of the runtime was used for training, with model reduction typically taking 1 to 2 hours and inference taking 2 to 4 hours for the larger datasets, including training the logistic regression readout.

\section{Per size statistical analysis}\label{section:per_size_stats}

In this section we provide statistical analysis and critical-difference diagrams per size of train dataset, for both image and tabular datasets. All per-size analyses use the same methodology as the aggregated analysis described in \S\ref{subsection:eval}: Friedman omnibus test with pairwise two-sided Wilcoxon signed-rank tests with Holm correction over all \({k}\choose{2}\) pairs as post-hoc, at \(\alpha = 0.05\). Critical-difference diagrams use the same procedure for clique formation.

\subsection{Image datasets}

We observe that the Friedman test does not reject the hypothesis that the ranks come from the same distribution for size \(n=100\) at \(\alpha=0.05\) (\(p=0.0770\)). The test rejects for all other sizes. We still report the post-hoc tables and critical-difference diagrams for illustrative purposes.

The main conclusion we can draw from the analysis is that AML is consistently in the clique with the best-ranked methods in the critical-difference diagram. Nonetheless, given the reduced number of datasets, in most cases there is only one clique, so no two methods can be distinguished by the Wilcoxon test. Note that, due to the inherent limitations of critical-difference diagrams as one-dimensional representations, if two methods are not distinguishable but can nonetheless be distinguished from a third method whose rank lies between them, a clique may not be drawn between the first two non-distinguishable methods. We observe this for the 200-example image setting: AML and MLP are not statistically distinguishable from each other (Table~\ref{table:images_all}), but the diagram cannot show this because Random forest, which is distinguishable from AML, shares its rank position.

\begin{figure}
\begin{center}
    \includegraphics[width=\textwidth]{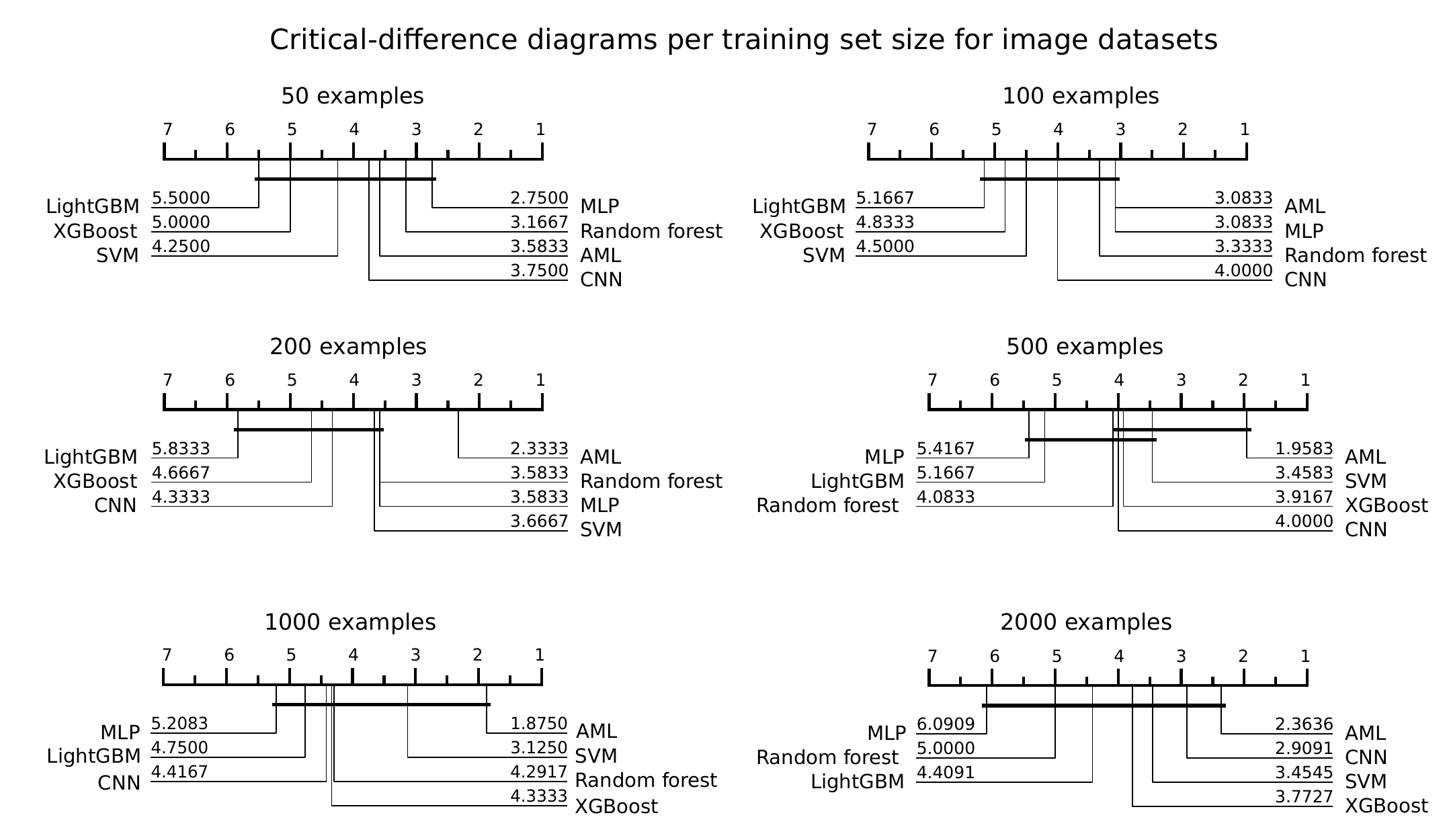}
    \caption{For image datasets, and per each of the considered training set sizes, critical-difference diagram including all analyzed methods, using the logistic regression readout as the AML classification variant.}
    \label{figure:critical_differences_images_all}
    \end{center}
\end{figure}

The per-size critical-difference diagrams can be found in Figure~\ref{figure:critical_differences_images_all}; and the tables for the post-hoc statistics can be found in Table~\ref{table:images_all} (sizes 50--200) and Table~\ref{table:images_all_cont} (sizes 500--2000).

\begin{table}
    \small
    \centering
    \caption{Per train set size, results of Friedman and post-hoc pairwise Wilcoxon signed-rank tests for AML compared with the baseline methods on image datasets, for sizes between 50 and 200 examples. Positive HL estimator values indicate AML achieves higher \(F_1\) than the baselines, and vice versa.}
    \label{table:images_all}
    \begin{tabular}{lllllll}
        \toprule
        \multirow{9}{*}{\begin{sideways}\parbox{2.2cm}{50 examples}\end{sideways}}
        & \multicolumn{6}{c}{Friedman test \(p\)-value: \(0.0208\)} \\[1pt]
        \cmidrule(l){2-7}\\[-10pt]
        & Method & Raw \(p\)-value & Adj.\ \(p\)-value & HL estimator & 95\% CI & Reject\\
        \cmidrule(l){2-7}\\[-10pt]
        & XGBoost & \(0.0342\) & \(0.5469\) & \(+0.0607\) & \([+0.0009, +0.1361]\) & No \\[2pt]
        &Random forest & \(0.9697\) & \(1.0000\) & \(-0.0016\) & \([-0.0228, +0.0136]\) & No \\[2pt]
        &LightGBM & \(0.0122\) & \(0.2197\) & \(+0.0602\) & \([+0.0106, +0.1405]\) & No \\[2pt]
        &SVM & \(0.5693\) & \(1.0000\) & \(+0.0121\) & \([-0.0378, +0.1521]\) & No \\[2pt]
        &MLP & \(0.7334\) & \(1.0000\) & \(-0.0040\) & \([-0.0381, +0.0236]\) & No \\[2pt]
        &CNN & \(0.2334\) & \(1.0000\) & \(+0.0241\) & \([-0.0165, +0.1259]\) & No \\[2pt]\hline\hline\\[-6pt]
        
        \multirow{9}{*}{\begin{sideways}\parbox{2.2cm}{100 examples}\end{sideways}}
        & \multicolumn{6}{c}{Friedman test \(p\)-value: \(0.0770\) (\textbf{does not reject})} \\[1pt]
        \cmidrule(l){2-7}\\[-10pt]
        & Method & Raw \(p\)-value & Adj.\ \(p\)-value & HL estimator & 95\% CI & Reject\\
        \cmidrule(l){2-7}\\[-10pt]
        & XGBoost & \(0.0269\) & \(0.5371\) & \(+0.0331\) & \([+0.0044, +0.1008]\) & No \\[2pt]
        &Random forest & \(0.8501\) & \(1.000\) & \(+0.0020\) & \([-0.0144, +0.0306]\) & No \\[2pt]
        &LightGBM & \(0.0161\) & \(0.3384\) & \(+0.0414\) & \([+0.0093, +0.1134]\) & No \\[2pt]
        &SVM & \(0.1514\) & \(1.0000\) & \(+0.0378\) & \([-0.0220, +0.0833]\) & No \\[2pt]
        &MLP & \(0.8501\) & \(1.0000\) & \(+0.0051\) & \([-0.0271, +0.0530]\) & No \\[2pt]
        &CNN & \(0.1294\) & \(1.0000\) & \(+0.0202\)& \([-0.0037, +0.1661]\) & No \\[2pt]\hline\hline\\[-6pt]

        \multirow{9}{*}{\begin{sideways}\parbox{2.2cm}{200 examples}\end{sideways}}
        & \multicolumn{6}{c}{Friedman test \(p\)-value: \(0.0053\)} \\[1pt]
        \cmidrule(l){2-7}\\[-10pt]
        & Method & Raw \(p\)-value & Adj.\ \(p\)-value & HL estimator & 95\% CI & Reject\\
        \cmidrule(l){2-7}\\[-10pt]
        & XGBoost & \(0.0024\) & \(0.0464\) & \(+0.0347\) & \([+0.0149, +0.0638]\) & Yes \\[2pt]
        &Random forest & \(<0.001\) & \(0.0205\) & \(+0.0145\) & \([+0.0051, +0.0297]\) & Yes \\[2pt]
        &LightGBM & \(0.0015\) & \(0.0293\) & \(+0.0464\) & \([+0.0223, +0.0907]\) & Yes \\[2pt]
        &SVM & \(0.0923\) & \(1.0000\) & \(+0.0206\) & \([-0.0044, +0.0545]\) & No \\[2pt]
        &MLP & \(0.4697\) & \(1.0000\)& \(+0.0096\) & \([-0.0190, +0.0574]\) & No \\[2pt]
        &CNN & \(0.2661\) & \(1.0000\) & \(+0.0293\) & \([-0.0210, +0.0796]\) & No \\[2pt]\hline\hline
    \end{tabular}
\end{table}

\begin{table}
    \small
    \centering
    \caption{Per train set size, results of Friedman and post-hoc pairwise Wilcoxon signed-rank tests for AML compared with the baseline methods on image datasets, for sizes between 500 and 2000 examples. Positive HL estimator values indicate AML achieves higher \(F_1\) than the baselines, and vice versa.}
    \label{table:images_all_cont}
    \begin{tabular}{lllllll}
        \hline\\[-8pt]
        \multirow{9}{*}{\begin{sideways}\parbox{2.3cm}{500 examples}\end{sideways}}
        & \multicolumn{6}{c}{Friedman test \(p\)-value: \(0.0026\)} \\[1pt]
        \cmidrule(l){2-7}\\[-10pt]
        & Method & Raw \(p\)-value & Adj.\ \(p\)-value & HL estimator & 95\% CI & Reject\\
        \cmidrule(l){2-7}\\[-10pt]
        & XGBoost & \(0.0269\) & \(0.4297\) & \(+0.0259\) & \([+0.0016, +0.0516]\) & No \\[2pt]
        &Random forest & \(<0.001\) & \(0.0103\) & \(+0.0217\) & \([+0.0112, +0.0375]\) & Yes \\[2pt]
        &LightGBM & \(0.0034\) & \(0.0615\) & \(+0.0344\) & \([+0.0138, +0.0612]\) & No \\[2pt]
        &SVM & \(0.2300\) & \(1.0000\) & \(+0.0235\) & \([-0.0069, +0.1572]\) & No \\[2pt]
        &MLP & \(<0.001\) & \(0.0103\)& \(+0.0489\)& \([+0.0148, +0.0849]\) & Yes \\[2pt]
        &CNN & \(0.0122\) & \(0.2075\) & \(+0.0579\) & \([+0.0038, +0.1109]\) & No \\[2pt]\hline\hline\\[-6pt]

        \multirow{9}{*}{\begin{sideways}\parbox{2.5cm}{1000 examples}\end{sideways}}
        & \multicolumn{6}{c}{Friedman test \(p\)-value: \(0.0027\)} \\[1pt]
        \cmidrule(l){2-7}\\[-10pt]
        & Method & Raw \(p\)-value & Adj.\ \(p\)-value & HL estimator & 95\% CI & Reject\\
        \cmidrule(l){2-7}\\[-10pt]
        & XGBoost & \(0.0210\) & \(0.3989\) & \(+0.0225\) & \([+0.0082, +0.0406]\) & No \\[2pt]
        &Random forest & \(0.0051\) & \(0.1020\) & \(+0.0241\) & \([+0.0126, +0.0353]\) & No \\[2pt]
        &LightGBM & \(0.0034\) & \(0.0718\) & \(+0.0246\) & \([+0.0093, +0.0410]\) & No \\[2pt]
        &SVM & \(0.1681\) & \(1.0000\) & \(+0.0115\) & \([-0.0090, +0.0570]\) & No \\[2pt]
        &MLP & \(0.0454\) & \(0.7726\)& \(+0.0430\) & \([+0.0039, +0.0765]\) &  No \\[2pt]
        &CNN & \(0.0342\) & \(0.6152\) & \(+0.0293\) & \([+0.0018, +0.0900]\) & No \\[2pt]\hline\hline\\[-6pt]

        \multirow{9}{*}{\begin{sideways}\parbox{2.5cm}{2000 examples}\end{sideways}}
        & \multicolumn{6}{c}{Friedman test \(p\)-value: \(<0.001\)} \\[1pt]
        \cmidrule(l){2-7}\\[-10pt]
        & Method & Raw \(p\)-value & Adj.\ \(p\)-value & HL estimator & 95\% CI & Reject\\
        \cmidrule(l){2-7}\\[-10pt]
        & XGBoost & \(0.0674\) & \(1.0000\) & \(+0.0124\) & \([-0.0063, +0.0313]\) & No \\[2pt]
        &Random forest & \(0.0137\) & \(0.2598\) & \(+0.0314\) & \([+0.0113, +0.0479]\) & No \\[2pt]
        &LightGBM & \(0.0674\) & \(1.0000\) & \(+0.0159\) & \([-0.0035, +0.0313]\) & No \\[2pt]
        &SVM & \(0.2061\) & \(1.0000\) & \(+0.0126\) & \([-0.0162, +0.0290]\) & No \\[2pt]
        &MLP & \(0.0322\) & \(0.5479\)& \(+0.0595\) & \([+0.0073, +0.1002]\)& No \\[2pt]
        &CNN & \(0.7002\) & \(1.0000\) & \(+0.0031\) & \([-0.0759, +0.0612]\) & No \\[2pt]\hline\hline\\[-6pt]
    \end{tabular}
\end{table}

\subsection{Tabular datasets}

We similarly observe that the Friedman test does not reject the hypothesis that the ranks come from the same distribution for one of the sizes at \(\alpha=0.05\), in this case for \(n=50\)  (\(p=0.0819\)).

\begin{figure}
\begin{center}
    \includegraphics[width=\textwidth]{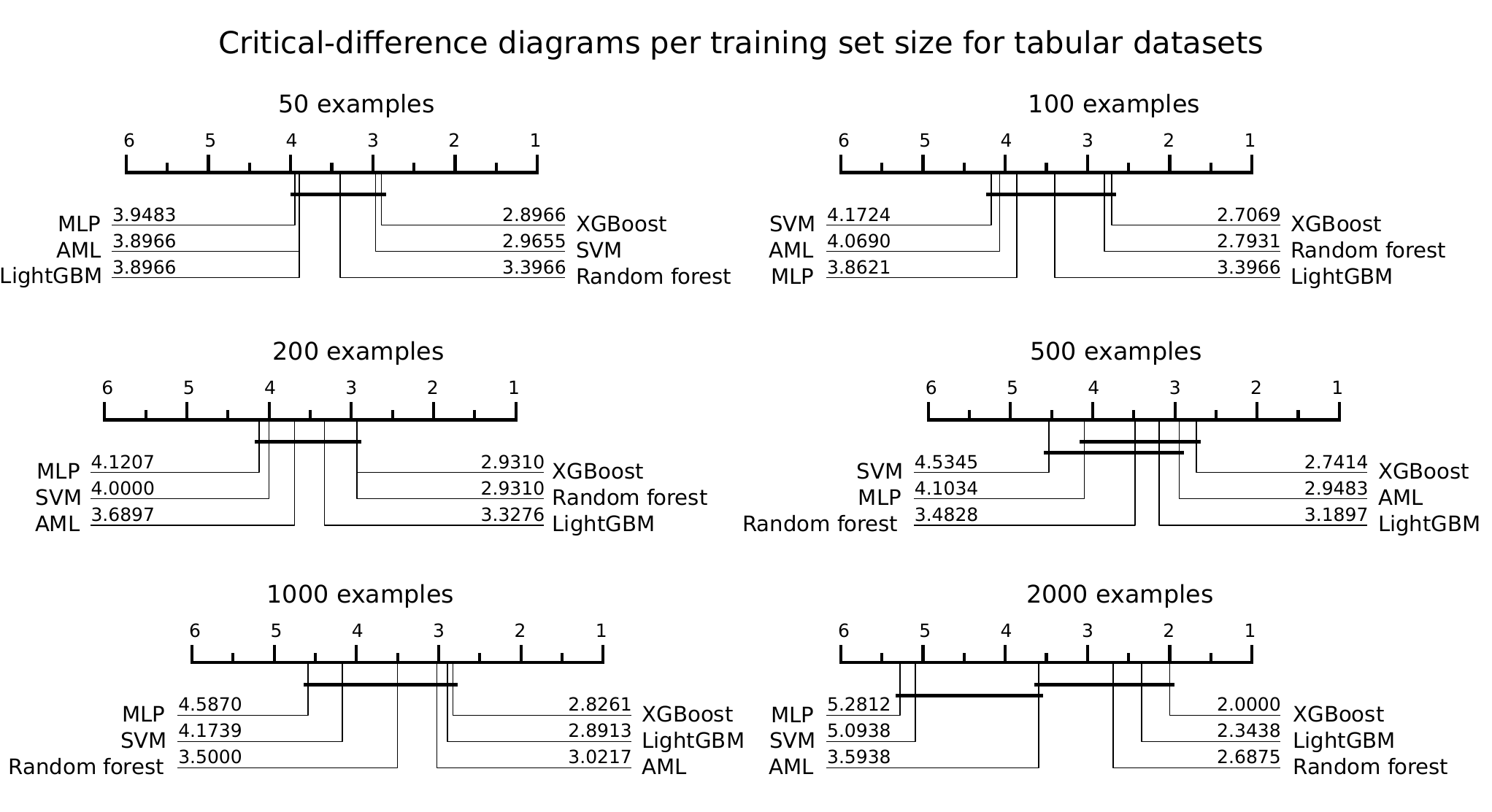}
    \caption{For tabular datasets, and per each of the considered training set sizes, critical-difference diagram including all analyzed methods, using the logistic regression readout as the AML classification variant.}
    \label{figure:critical_differences_tabular_all}
\end{center}
\end{figure}

We observe in this case that XGBoost is consistently the algorithm with the best mean rank, with AML not being distinguishable from XGBoost in any size. In fact, AML is not distinguishable from any method at any individual training-set size, even though the aggregated analysis (Table \ref{table:tabular_p_values}) distinguishes AML from XGBoost. This reflects the lower statistical power of per-size analyses compared to aggregated analysis. The per-size critical-difference diagrams can be found in Figure~\ref{figure:critical_differences_tabular_all}; and the tables for the post-hoc statistics can be found in Table~\ref{table:tabular_all}.

\begin{table}
    \small
    \centering
    \caption{Per train set size, results of Friedman and post-hoc pairwise Wilcoxon signed-rank tests for AML compared with the baseline methods on tabular datasets. Positive HL estimator values indicate AML achieves higher \(F_1\) than the baselines, and vice versa.}
    \label{table:tabular_all}
    \begin{tabular}{lllllll}
        \toprule
        \multirow{8}{*}{\begin{sideways}\parbox{2cm}{50 examples}\end{sideways}}
        & \multicolumn{6}{c}{Friedman test \(p\)-value: \(0.0819\) (\textbf{does not reject})} \\[1pt]
        \cmidrule(l){2-7}\\[-10pt]
        & Method & Raw \(p\)-value & Adj.\ \(p\)-value & HL estimator & 95\% CI & Reject\\
        \cmidrule(l){2-7}\\[-10pt]
        & XGBoost & \(0.1056\) & \(1.0000\) & \(-0.0200\) & \([-0.0415, +0.0048]\) & No \\[2pt]
        &Random forest & \(0.5221\) & \(1.0000\) & \(-0.0090\) & \([-0.0463, +0.0210]\) & No \\[2pt]
        &LightGBM & \(0.5504\) & \(1.0000\) & \(+0.0129\) & \([-0.0199, +0.0454]\) & No \\[2pt]
        &SVM & \(0.3467\) & \(1.0000\) & \(-0.0175\) & \([-0.0555, +0.0250]\) & No \\[2pt]
        &MLP & \(0.8382\) & \(1.0000\) & \(+0.0038\) & \([-0.0368, +0.0531]\) & No\\[2pt]\hline\hline\\[-6pt]
        
        \multirow{9}{*}{\begin{sideways}\parbox{1.8cm}{100 examples}\end{sideways}}
        & \multicolumn{6}{c}{Friedman test \(p\)-value: \(0.0044\)} \\[1pt]
        \cmidrule(l){2-7}\\[-10pt]
        & Method & Raw \(p\)-value & Adj.\ \(p\)-value & HL estimator & 95\% CI & Reject\\
        \cmidrule(l){2-7}\\[-10pt]
        & XGBoost & \(0.0065\) & \(0.0968\) & \(-0.0337\) & \([-0.0563, -0.0104]\) & No \\[2pt]
        &Random forest & \(0.0111\) & \(0.1556\) & \(-0.0299\) & \([-0.0536, -0.0062]\) & No \\[2pt]
        &LightGBM & \(0.1035\) & \(0.8279\) & \(-0.0114\) &\([-0.0280, +0.0021]\) & No \\[2pt]
        &SVM & \(0.8480\) & \(1.0000\) & \(-0.0017\) & \([-0.0329, +0.0273]\) & No \\[2pt]
        &MLP & \(0.7172\) & \(1.0000\)& \(-0.0058\) & \([-0.0334, +0.0253]\) & No \\[2pt]\hline\hline\\[-6pt]

        \multirow{9}{*}{\begin{sideways}\parbox{1.8cm}{200 examples}\end{sideways}}
        & \multicolumn{6}{c}{Friedman test \(p\)-value: \(0.0472\)} \\[1pt]
        \cmidrule(l){2-7}\\[-10pt]
        & Method & Raw \(p\)-value & Adj.\ \(p\)-value & HL estimator & 95\% CI & Reject\\
        \cmidrule(l){2-7}\\[-10pt]
        & XGBoost & \(0.0543\) & \(0.6939\) & \(-0.0167\) & \([-0.0332, +0.0009]\) & No \\[2pt]
        &Random forest & \(0.1683\) & \(1.0000\) & \(-0.0079\) & \([-0.0224, +0.0050]\) & No \\[2pt]
        &LightGBM & \(0.4047\) & \(1.0000\) & \(-0.0060\) & \([-0.0206, +0.0092]\) & No \\[2pt]
        &SVM & \(0.4420\) & \(1.0000\) & \(+0.0102\) & \([-0.0135, +0.0323]\)& No \\[2pt]
        &MLP & \(0.1557\) & \(1.0000\)& \(+0.0139\) & \([-0.0059, +0.0373]\) & No \\[2pt]\hline\hline\\[-6pt]

        \multirow{9}{*}{\begin{sideways}\parbox{1.8cm}{500 examples}\end{sideways}}
        & \multicolumn{6}{c}{Friedman test \(p\)-value: \(0.0012\)} \\[1pt]
        \cmidrule(l){2-7}\\[-10pt]
        & Method & Raw \(p\)-value & Adj.\ \(p\)-value & HL estimator & 95\% CI & Reject\\
        \cmidrule(l){2-7}\\[-10pt]
        & XGBoost & \(0.3467\) & \(1.0000\) & \(-0.0093\) & \([-0.0301, +0.0068]\) & No \\[2pt]
        &Random forest & \(0.4593\) & \(1.0000\) & \(+0.0028\) & \([-0.0055, +0.0138]\) & No \\[2pt]
        &LightGBM & \(0.6391\) & \(1.0000\) & \(-0.0044\) & \([-0.0228, +0.0129]\) & No \\[2pt]
        &SVM & \(0.0052\) & \(0.0739\) & \(+0.0259\) & \([+0.0073, +0.0453]\) & No \\[2pt]
        &MLP & \(0.1982\) & \(1.0000\)& \(+0.0120\)& \([-0.0073, +0.0404]\) & No \\[2pt]\hline\hline\\[-6pt]

        \multirow{9}{*}{\begin{sideways}\parbox{2cm}{1000 examples}\end{sideways}}
        & \multicolumn{6}{c}{Friedman test \(p\)-value: \(0.0033\)} \\[1pt]
        \cmidrule(l){2-7}\\[-10pt]
        & Method & Raw \(p\)-value & Adj.\ \(p\)-value & HL estimator & 95\% CI & Reject\\
        \cmidrule(l){2-7}\\[-10pt]
        & XGBoost & \(0.8967\) & \(1.0000\) & \(+0.0011\) & \([-0.0277, +0.0158]\) & No \\[2pt]
        &Random forest & \(0.5399\) & \(1.0000\) & \(+0.0061\) & \([-0.0126, +0.0327]\) & No \\[2pt]
        &LightGBM & \(0.5600\) & \(1.0000\) & \(+0.0059\) & \([-0.0107, +0.0265]\) & No \\[2pt]
        &SVM & \(0.0149\) & \(0.1784\) & \(+0.0194\) & \([+0.0040, +0.0458]\) & No \\[2pt]
        &MLP & \(0.0049\) & \(0.0679\)& \(+0.0252\) & \([+0.0067, +0.0611]\) & No \\[2pt]\hline\hline\\[-6pt]

        \multirow{9}{*}{\begin{sideways}\parbox{2cm}{2000 examples}\end{sideways}}
        & \multicolumn{6}{c}{Friedman test \(p\)-value: \(<0.001\)} \\[1pt]
        \cmidrule(l){2-7}\\[-10pt]
        & Method & Raw \(p\)-value & Adj.\ \(p\)-value & HL estimator & 95\% CI & Reject\\
        \cmidrule(l){2-7}\\[-10pt]
        & XGBoost & \(0.0110\) & \(0.0879\) & \(-0.0164\) & \([-0.0751, -0.0046]\) & No \\[2pt]
        &Random forest & \(0.2744\) & \(0.9772\) & \(-0.0091\)& \([-0.0314, +0.0076]\) & No \\[2pt]
        &LightGBM & \(0.0063\) & \(0.0566\) & \(-0.0123\) & \([-0.0567, -0.0033]\) & No \\[2pt]
        &SVM & \(0.0156\) & \(0.1085\) & \(+0.0256\) & \([+0.0070, +0.0592]\) & No \\[2pt]
        &MLP & \(0.0214\) & \(0.1286\)& \(+0.0387\) & \([+0.0090, +0.0677]\) & No \\[2pt]\hline\hline\\[-6pt]
    \end{tabular}
\end{table}

\section{Extended experimental results}\label{section:extended_results}

In Table~\ref{table:results_image_datasets} we provide accuracies and \(F_1\) scores for all image datasets, for the single AML run as well as for the models selected after hyperparameter selection for all the parametrized baselines. We provide the same information for tabular datasets in Table~\ref{table:results_tabular_datasets}. Note that not all tabular datasets are large enough to reserve 1000 or 2000 training examples, and similarly only 1000 examples were reserved for training in COIL-20.

\begin{landscape}
\begin{table}
    \small
    \centering
    \caption{Experimental results in accuracy and macro-\(F_1\) for image datasets, for AML in both classification variants and baselines. The headline results for AML use the logistic regression readout; results obtained via fewest misses are also reported.}
    \label{table:results_image_datasets}
    \begin{tabular}{cccccccccccccccccc}
    \toprule
    &\multirow{2}{*}{\# train} & \multicolumn{2}{c}{AML} & \multicolumn{2}{c}{Fewest misses} &\multicolumn{2}{c}{XGBoost} & \multicolumn{2}{c}{LightGBM} & \multicolumn{2}{c}{Random Forest} & \multicolumn{2}{c}{SVM} & \multicolumn{2}{c}{MLP} & \multicolumn{2}{c}{CNN} \\
        && acc.\  & \(F_1\) & acc.\ & \(F_1\) & acc.\ & \(F_1\) & acc.\ & \(F_1\)  & acc.\ & \(F_1\) & acc.\ & \(F_1\) & acc.\ & \(F_1\) & acc.\ & \(F_1\)\\[2pt] \hline\midrule \\[-7pt]
    
    \multirow{4}{*}{\begin{sideways}\parbox{1.6cm}{MNIST}\end{sideways}}
            
    & 50 & 60.74\% & 0.5903 & 57.37\% & 0.5537 & 43.99\% & 0.4263 & 51.22\% & 0.5001 & 61.45\% & 0.5955 & 64.22\% & 0.6298 & 63.32\% & 0.6220 & 66.37\% & 0.6537 \\[2pt]
    & 100 & 66.72\% & 0.6436 & 64.07\% & 0.6081 & 62.08\% & 0.6058 & 60.46\% & 0.5872 & 68.44\% & 0.6614 & 68.42\% & 0.6675 & 69.87\% & 0.6849 & 67.66\% & 0.6578 \\[2pt]
    & 200 & 75.75\% & 0.7433 & 74.16\% & 0.7240 & 71.49\% & 0.7063 & 70.15\% & 0.6934 & 74.09\% & 0.7250 & 77.59\% & 0.7651 & 77.27\% & 0.7623 & 78.86\% & 0.7835 \\[2pt]
    & 500 & 86.76\% & 0.8655 & 83.67\% & 0.8339 & 82.28\% & 0.8197 & 80.77\% & 0.8044 & 83.77\% & 0.8334 & 87.57\% & 0.8744 & 86.01\% & 0.8580 & 86.41\% & 0.8624 \\[2pt]
    & 1000 & 90.74\% & 0.9061 & 88.79\% & 0.8863 & 87.82\% & 0.8767 & 87.58\% & 0.8740 & 88.60\% & 0.8841 & 91.30\% & 0.9118 & 89.34\% & 0.8916 & 90.57\% & 0.9044 \\[2pt]
    & 2000 & 92.99\% & 0.9290 & 91.43\% & 0.9130 & 90.91\% & 0.9081 & 91.34\% & 0.9124 & 91.76\% & 0.9164 & 93.72\% & 0.9363 & 89.40\% & 0.8921 & 93.75\% & 0.9364 \\[2pt]\hline\\[-6pt]

    \multirow{4}{*}{\begin{sideways}\parbox{2.2cm}{Fashion-MNIST}\end{sideways}}
    & 50 & 62.65\% & 0.6187 & 59.17\% & 0.5806 & 51.29\% & 0.5020 & 48.88\% & 0.4736 & 62.92\% & 0.6275 & 66.00\% & 0.6605 & 67.83\% & 0.6748 & 64.67\% & 0.6479 \\[2pt]
    & 100 & 69.12\% & 0.6929 & 66.15\% & 0.6597 & 62.66\% & 0.6289 & 63.87\% & 0.6419 & 70.34\% & 0.7040 & 57.65\% & 0.6069 & 71.63\% & 0.7195 & 65.33\% & 0.6561 \\[2pt]
    & 200 & 74.74\% & 0.7471 & 73.05\% & 0.7328 & 70.77\% & 0.7108 & 70.64\% & 0.7096 & 74.61\% & 0.7428 & 75.58\% & 0.7573 & 75.07\% & 0.7517 & 67.58\% & 0.6514 \\[2pt]
    & 500 & 79.14\% & 0.7892 & 77.60\% & 0.7739 & 76.99\% & 0.7705 & 76.71\% & 0.7664 & 78.13\% & 0.7767 & 78.96\% & 0.7890 & 78.81\% & 0.7888 & 79.22\% & 0.7916 \\[2pt]
    & 1000 & 81.74\% & 0.8162 & 79.69\% & 0.7964 & 80.43\% & 0.8048 & 79.65\% & 0.7964 & 79.78\% & 0.7946 & 81.25\% & 0.8132 & 76.63\% & 0.7687 & 79.67\% & 0.7946 \\[2pt]
    & 2000 & 83.78\% & 0.8362 & 81.63\% & 0.8147 & 82.85\% & 0.8281 & 82.85\% & 0.8281 & 82.39\% & 0.8211 & 82.85\% & 0.8289 & 81.58\% & 0.8170 & 83.08\% & 0.8319 \\[2pt]
    \hline\\[-6pt]

    \multirow{4}{*}{\begin{sideways}\parbox{1.85cm}{CIFAR-10}\end{sideways}}
    & 50 & 18.00\% & 0.1571 & 16.79\% & 0.1450 & 15.46\% & 0.1532 & 15.33\% & 0.1464 & 20.76\% & 0.1858 & 19.09\% & 0.1657 & 20.18\% & 0.1805 & 19.53\% & 0.1736 \\[2pt]
    & 100 & 21.31\% & 0.1849 & 20.26\% & 0.1614 & 19.10\% & 0.1787 & 19.22\% & 0.1840 & 22.04\% & 0.2017 & 17.44\% & 0.1145 & 21.99\% & 0.2029 & 22.05\% & 0.2062 \\[2pt]
    & 200 & 27.61\% & 0.2671 & 26.34\% & 0.2432 & 24.12\% & 0.2355 & 23.10\% & 0.2250 & 27.14\% & 0.2620 & 26.32\% & 0.2477 & 25.33\% & 0.2416 & 25.35\% & 0.2417 \\[2pt]
    & 500 & 34.68\% & 0.3384 & 33.12\% & 0.3164 & 31.65\% & 0.3066 & 30.40\% & 0.2960 & 32.73\% & 0.3129 & 30.14\% & 0.2981 & 28.40\% & 0.2699 & 31.71\% & 0.3016 \\[2pt]
    & 1000 & 38.01\% & 0.3748 & 36.39\% & 0.3554 & 35.57\% & 0.3508 & 34.35\% & 0.3386 & 35.64\% & 0.3467 & 36.46\% & 0.3634 & 31.87\% & 0.3113 & 39.12\% & 0.3922 \\[2pt]
    & 2000 & 42.58\% & 0.4223 & 39.49\% & 0.3869 & 40.30\% & 0.3996 & 39.31\% & 0.3911 & 38.48\% & 0.3746 & 40.53\% & 0.4052 & 36.08\% & 0.3577 & 44.29\% & 0.4420 \\[2pt]
    \hline\\[-6pt]

    \multirow{4}{*}{\begin{sideways}\parbox{2.25cm}{OrganCMNIST}\end{sideways}}
    & 50 & 54.53\% & 0.4544 & 47.59\% & 0.3830 & 34.63\% & 0.2640 & 32.92\% & 0.2621 & 53.71\% & 0.4403 & 22.33\% & 0.0332 & 44.97\% & 0.3867 & 46.68\% & 0.4002 \\[2pt]
    & 100 & 60.35\% & 0.5273 & 56.58\% & 0.4904 & 47.77\% & 0.4191 & 46.68\% & 0.4018 & 56.89\% & 0.4973 & 49.04\% & 0.4363 & 51.89\% & 0.4542 & 57.08\% & 0.5133 \\[2pt]
    & 200 & 71.41\% & 0.6953 & 66.15\% & 0.6248 & 63.90\% & 0.6185 & 61.09\% & 0.5868 & 69.06\% & 0.6668 & 63.05\% & 0.6160 & 60.11\% & 0.5803 & 60.56\% & 0.5804 \\[2pt]
    & 500 & 76.22\% & 0.7477 & 71.86\% & 0.6999 & 71.84\% & 0.7016 & 71.25\% & 0.6980 & 73.04\% & 0.7141 & 70.24\% & 0.6927 & 63.44\% & 0.6173 & 68.61\% & 0.6298 \\[2pt]
    & 1000 & 79.99\% & 0.7850 & 75.47\% & 0.7356 & 74.74\% & 0.7333 & 75.49\% & 0.7392 & 76.76\% & 0.7496 & 74.31\% & 0.7281 & 67.76\% & 0.6559 & 76.36\% & 0.7408 \\[2pt]
    & 2000 & 82.45\% & 0.8102 & 78.42\% & 0.7709 & 79.35\% & 0.7788 & 80.26\% & 0.7875 & 78.60\% & 0.7685 & 76.79\% & 0.7539 & 69.07\% & 0.6700 & 80.79\% & 0.7830 \\
\bottomrule
\end{tabular}
\end{table}

\begin{table}
    \small
    \centering
    \begin{tabular}{cccccccccccccccccc}
        \toprule
    &\multirow{2}{*}{\# train} & \multicolumn{2}{c}{AML} & \multicolumn{2}{c}{Fewest misses} &\multicolumn{2}{c}{XGBoost} & \multicolumn{2}{c}{LightGBM} & \multicolumn{2}{c}{Random Forest} & \multicolumn{2}{c}{SVM} & \multicolumn{2}{c}{MLP} & \multicolumn{2}{c}{CNN} \\
        && acc.\ & \(F_1\) & acc.\ & \(F_1\) & acc.\ & \(F_1\) & acc.\ & \(F_1\)  & acc.\ & \(F_1\) & acc.\ & \(F_1\) & acc.\ & \(F_1\) & acc.\ & \(F_1\)\\[2pt] \hline\midrule \\[-7pt]
    
    \multirow{4}{*}{\begin{sideways}\parbox{2.1cm}{BloodMNIST}\end{sideways}}

    & 50 & 51.33\% & 0.4613 & 48.52\% & 0.4057 & 46.80\% & 0.4251 & 44.90\% & 0.4065 & 53.99\% & 0.4733 & 46.62\% & 0.4179 & 49.17\% & 0.4468 & 51.68\% & 0.4868 \\[2pt]
    & 100 & 65.21\% & 0.5897 & 63.81\% & 0.5440 & 59.78\% & 0.5652 & 59.89\% & 0.5438 & 66.71\% & 0.6134 & 63.46\% & 0.5860 & 54.49\% & 0.4780 & 51.86\% & 0.4210 \\[2pt]
    & 200 & 72.03\% & 0.6399 & 67.85\% & 0.5513 & 69.10\% & 0.6199 & 66.30\% & 0.5837 & 67.49\% & 0.5916 & 67.20\% & 0.6213 & 61.03\% & 0.5167 & 63.72\% & 0.5158 \\[2pt]
    & 500 & 79.63\% & 0.7581 & 75.36\% & 0.7024 & 77.52\% & 0.7381 & 77.02\% & 0.7295 & 74.45\% & 0.6836 & 74.31\% & 0.7066 & 64.25\% & 0.6169 & 68.14\% & 0.6450 \\[2pt]
    & 1000 & 81.29\% & 0.7796 & 76.79\% & 0.7171 & 80.01\% & 0.7714 & 79.04\% & 0.7581 & 76.82\% & 0.7377 & 77.40\% & 0.7430 & 74.74\% & 0.7172 & 74.04\% & 0.6709 \\[2pt]
    & 2000 & 85.12\% & 0.8311 & 80.06\% & 0.7639 & 83.89\% & 0.8188 & 83.72\% & 0.8143 & 80.24\% & 0.7772 & 82.52\% & 0.8021 & 71.56\% & 0.6953 & 81.53\% & 0.7684 \\[2pt]
    \hline\\[-6pt]

    \multirow{4}{*}{\begin{sideways}\parbox{2.15cm}{DermaMNIST}\end{sideways}}
    & 50 & 59.90\% & 0.1939 & 63.84\% & 0.1780 & 54.16\% & 0.1965 & 54.36\% & 0.1943 & 60.00\% & 0.1809 & 46.63\% & 0.1675 & 54.66\% & 0.1940 & 58.60\% & 0.1879 \\[2pt]
    & 100 & 66.18\% & 0.2344 & 66.33\% & 0.1803 & 59.05\% & 0.2224 & 60.50\% & 0.2508 & 65.99\% & 0.2033 & 66.88\% & 0.1145 & 58.15\% & 0.2089 & 52.77\% & 0.2229 \\[2pt]
    & 200 & 66.48\% & 0.2453 & 67.43\% & 0.2054 & 62.19\% & 0.2535 & 64.49\% & 0.2528 & 67.28\% & 0.1911 & 66.88\% & 0.1145 & 60.30\% & 0.2456 & 62.49\% & 0.2103 \\[2pt]
    & 500 & 68.58\% & 0.2716 & 68.18\% & 0.2227 & 56.16\% & 0.3047 & 52.17\% & 0.2883 & 68.18\% & 0.2291 & 66.88\% & 0.1145 & 65.99\% & 0.1749 & 66.93\% & 0.1354 \\[2pt]
    & 1000 & 69.18\% & 0.2862 & 69.63\% & 0.2549 & 62.34\% & 0.3293 & 54.56\% & 0.2997 & 69.33\% & 0.2373 & 67.53\% & 0.1549 & 65.49\% & 0.2656 & 69.53\% & 0.2149 \\[2pt]
    & 2000 & 70.47\% & 0.3124 & 70.47\% & 0.2609 & 61.75\% & 0.3830 & 57.41\% & 0.3507 & 69.68\% & 0.2442 & 69.63\% & 0.2770 & 67.88\% & 0.1919 & 67.28\% & 0.1317 \\[2pt]
    \hline\\[-6pt]

    \multirow{4}{*}{\begin{sideways}\parbox{2.4cm}{PneumoniaMNIST}\end{sideways}}
    & 50 & 74.04\% & 0.6726 & 71.31\% & 0.6081 & 75.32\% & 0.7069 & 71.31\% & 0.6500 & 72.92\% & 0.6485 & 76.12\% & 0.7087 & 80.93\% & 0.7796 & 62.50\% & 0.3846 \\[2pt]
    & 100 & 80.77\% & 0.7709 & 78.37\% & 0.7324 & 83.17\% & 0.8100 & 80.45\% & 0.7744 & 75.00\% & 0.6858 & 84.29\% & 0.8297 & 74.20\% & 0.6582 & 62.82\% & 0.3939 \\[2pt]
    & 200 & 78.53\% & 0.7376 & 77.40\% & 0.7166 & 76.44\% & 0.7207 & 75.16\% & 0.7137 & 77.88\% & 0.7260 & 77.72\% & 0.7245 & 83.17\% & 0.8081 & 82.85\% & 0.8049 \\[2pt]
    & 500 & 83.49\% & 0.8077 & 82.53\% & 0.7922 & 84.78\% & 0.8245 & 83.49\% & 0.8082 & 83.49\% & 0.8071 & 83.81\% & 0.8124 & 79.49\% & 0.7485 & 77.08\% & 0.7037 \\[2pt]
    & 1000 & 84.29\% & 0.8162 & 82.53\% & 0.7916 & 83.33\% & 0.8066 & 83.17\% & 0.8071 & 82.85\% & 0.8002 & 83.81\% & 0.8086 & 81.89\% & 0.7805 & 81.89\% & 0.7783 \\[2pt]
    & 2000 & 85.26\% & 0.8280 & 83.97\% & 0.8108 & 83.97\% & 0.8135 & 83.97\% & 0.8135 & 84.13\% & 0.8167 & 85.58\% & 0.8322 & 80.93\% & 0.7642 & 83.97\% & 0.8073 \\[2pt]
    \hline\\[-6pt]

    \multirow{4}{*}{\begin{sideways}\parbox{2.4cm}{\centering Aerial Cactus\\Identification}\end{sideways}}
    & 50 & 79.93\% & 0.6279 & 79.32\% & 0.5863 & 77.74\% & 0.6696 & 75.88\% & 0.6663 & 82.80\% & 0.7010 & 87.44\% & 0.8034 & 79.28\% & 0.5970 & 75.53\% & 0.4303 \\[2pt]
    & 100 & 83.50\% & 0.7175 & 82.91\% & 0.6948 & 80.16\% & 0.7206 & 78.72\% & 0.6976 & 83.75\% & 0.7107 & 89.46\% & 0.8475 & 83.26\% & 0.7720 & 78.73\% & 0.5539 \\[2pt]
    & 200 & 86.84\% & 0.7851 & 85.32\% & 0.7458 & 86.21\% & 0.7916 & 85.22\% & 0.7779 & 86.65\% & 0.7827 & 89.89\% & 0.8523 & 86.95\% & 0.8195 & 93.68\% & 0.9113 \\[2pt]
    & 500 & 90.77\% & 0.8602 & 88.78\% & 0.8194 & 90.56\% & 0.8665 & 89.88\% & 0.8531 & 90.31\% & 0.8574 & 91.26\% & 0.8711 & 88.49\% & 0.8415 & 93.95\% & 0.9185 \\[2pt]
    & 1000 & 91.67\% & 0.8787 & 90.57\% & 0.8557 & 89.65\% & 0.8563 & 91.14\% & 0.8752 & 90.85\% & 0.8661 & 92.08\% & 0.8879 & 85.91\% & 0.8225 & 94.67\% & 0.9258 \\[2pt]
    & 2000 & 92.43\% & 0.8913 & 91.41\% & 0.8706 & 92.49\% & 0.8976 & 92.28\% & 0.8954 & 90.65\% & 0.8664 & 92.60\% & 0.8947 & 88.85\% & 0.8456 & 96.29\% & 0.9497 \\
    \bottomrule

    \end{tabular}
\end{table}

\begin{table}
\small
    \centering
    \begin{tabular}{cccccccccccccccccc}
        \toprule
    &\multirow{2}{*}{\# train} & \multicolumn{2}{c}{AML} & \multicolumn{2}{c}{Fewest misses} &\multicolumn{2}{c}{XGBoost} & \multicolumn{2}{c}{LightGBM} & \multicolumn{2}{c}{Random Forest} & \multicolumn{2}{c}{SVM} & \multicolumn{2}{c}{MLP} & \multicolumn{2}{c}{CNN} \\
        && acc.\ & \(F_1\) & acc.\ & \(F_1\) & acc.\ & \(F_1\) & acc.\ & \(F_1\) & acc.\ & \(F_1\)  & acc.\ & \(F_1\) & acc.\ & \(F_1\) & acc.\ & \(F_1\)\\[2pt] \hline\midrule \\[-7pt]

    \multirow{4}{*}{\begin{sideways}\parbox{2.2cm}{\centering Street View\\House Number}\end{sideways}}
    & 50 & 13.66\% & 0.1024 & 12.93\% & 0.0900 & 12.14\% & 0.1015 & 12.81\% & 0.1048 & 14.62\% & 0.1194 & 15.17\% & 0.1047 & 15.67\% & 0.1240 & 15.94\% & 0.0275 \\[2pt]
    & 100 & 17.27\% & 0.1226 & 17.03\% & 0.1002 & 14.74\% & 0.1235 & 14.52\% & 0.1240 & 18.05\% & 0.1372 & 15.94\% & 0.0275 & 20.18\% & 0.1618 & 20.58\% & 0.1334 \\[2pt]
    & 200 & 20.27\% & 0.1552 & 19.67\% & 0.1256 & 16.29\% & 0.1354 & 17.04\% & 0.1368 & 20.04\% & 0.1416 & 19.83\% & 0.1149 & 21.37\% & 0.1638 & 18.90\% & 0.0637 \\[2pt]
    & 500 & 28.96\% & 0.2189 & 25.80\% & 0.1722 & 22.94\% & 0.1868 & 24.08\% & 0.1908 & 27.61\% & 0.2007 & 32.00\% & 0.2714 & 29.14\% & 0.2080 & 19.59\% & 0.0328 \\[2pt]
    & 1000 & 35.96\% & 0.3038 & 30.66\% & 0.2410 & 30.88\% & 0.2654 & 32.51\% & 0.2745 & 35.93\% & 0.3097 & 41.42\% & 0.3785 & 45.78\% & 0.4250 & 19.46\% & 0.0330 \\[2pt]
    & 2000 & 43.70\% & 0.3907 & 38.66\% & 0.3219 & 43.67\% & 0.3904 & 43.82\% & 0.3877 & 47.06\% & 0.4224 & 51.49\% & 0.4793 & 53.36\% & 0.5119 & 75.03\% & 0.7232 \\[2pt]
    \hline\\[-6pt]

    \multirow{4}{*}{\begin{sideways}\parbox{2.3cm}{\centering STL-10}\end{sideways}}
    & 50 & 22.45\% & 0.2179 & 20.31\% & 0.1873 & 17.81\% & 0.1700 & 17.62\% & 0.1680 & 22.49\% & 0.1977 & 10.00\% & 0.0182 & 22.60\% & 0.2003 & 19.65\% & 0.1770 \\[2pt]
    & 100 & 23.95\% & 0.2181 & 23.42\% & 0.2047 & 20.50\% & 0.1947 & 20.51\% & 0.1938 & 25.11\% & 0.2290 & 18.54\% & 0.1466 & 23.81\% & 0.2177 & 22.24\% & 0.1937 \\[2pt]
    & 200 & 30.16\% & 0.2945 & 29.28\% & 0.2826 & 26.34\% & 0.2560 & 26.12\% & 0.2551 & 29.91\% & 0.2893 & 28.58\% & 0.2708 & 27.94\% & 0.2777 & 26.16\% & 0.2304 \\[2pt]
    & 500 & 35.09\% & 0.3379 & 33.44\% & 0.3169 & 32.57\% & 0.3178 & 31.51\% & 0.3100 & 33.36\% & 0.3201 & 33.64\% & 0.3304 & 30.40\% & 0.2984 & 33.71\% & 0.3313 \\[2pt]
    & 1000 & 39.14\% & 0.3866 & 36.64\% & 0.3591 & 36.49\% & 0.3641 & 35.33\% & 0.3545 & 35.83\% & 0.3542 & 36.31\% & 0.3639 & 28.88\% & 0.2897 & 36.81\% & 0.3666 \\[2pt]
    & 2000 & 42.94\% & 0.4254 & 39.99\% & 0.3909 & 41.40\% & 0.4131 & 39.56\% & 0.3959 & 38.74\% & 0.3753 & 40.79\% & 0.4022 & 36.94\% & 0.3604 & 41.92\% & 0.4047 \\[2pt]
    \hline\\[-6pt]

    \multirow{4}{*}{\begin{sideways}\parbox{2.4cm}{\centering Kuzushiji-49}\end{sideways}}
    & 50 & 18.72\% & 0.1252 & 15.08\% & 0.0911 & 6.92\% & 0.0465 & 7.09\% & 0.0469 & 16.15\% & 0.1000 & 17.66\% & 0.1163 & 19.90\% & 0.1358 & 19.91\% & 0.1362 \\[2pt]
    & 100 & 22.78\% & 0.1710 & 16.15\% & 0.1290 & 9.71\% & 0.0703 & 9.64\% & 0.0697 & 21.84\% & 0.1692 & 16.82\% & 0.1205 & 22.76\% & 0.1767 & 24.18\% & 0.1839 \\[2pt]
    & 200 & 32.58\% & 0.2650 & 25.06\% & 0.2037 & 16.61\% & 0.1309 & 16.06\% & 0.1256 & 32.80\% & 0.2667 & 28.10\% & 0.2178 & 28.75\% & 0.2438 & 31.73\% & 0.2607 \\[2pt]
    & 500 & 42.87\% & 0.3742 & 36.22\% & 0.2909 & 29.67\% & 0.2461 & 28.56\% & 0.2351 & 40.71\% & 0.3523 & 9.75\% & 0.0372 & 37.94\% & 0.3282 & 42.33\% & 0.3691 \\[2pt]
    & 1000 & 48.71\% & 0.4357 & 42.49\% & 0.3565 & 38.17\% & 0.3244 & 35.69\% & 0.3058 & 48.19\% & 0.4292 & 48.40\% & 0.4295 & 45.91\% & 0.4124 & 44.47\% & 0.4088 \\[2pt]
    & 2000 & 55.41\% & 0.5117 & 48.67\% & 0.4250 & 48.05\% & 0.4287 & 47.22\% & 0.4202 & 53.45\% & 0.4841 & 54.05\% & 0.4934 & 54.30\% & 0.5050 & 57.05\% & 0.5277 \\[2pt]
    \hline\\[-6pt]

    \multirow{4}{*}{\begin{sideways}\parbox{2cm}{\centering COIL-20}\end{sideways}}
    & 50 & 69.32\% & 0.6606 & 58.41\% & 0.5541 & 44.77\% & 0.3893 & 43.18\% & 0.3792 & 69.55\% & 0.6623 & 52.95\% & 0.5085 & 64.77\% & 0.6133 & 65.00\% & 0.6143 \\[2pt]
    & 100 & 84.09\% & 0.8413 & 78.18\% & 0.7570 & 63.64\% & 0.6210 & 62.05\% & 0.5926 & 83.64\% & 0.8262 & 83.41\% & 0.8357 & 82.95\% & 0.8285 & 83.86\% & 0.8357 \\[2pt]
    & 200 & 94.77\% & 0.9468 & 90.91\% & 0.9042 & 87.73\% & 0.8679 & 81.36\% & 0.8009 & 92.50\% & 0.9237 & 92.95\% & 0.9267 & 91.36\% & 0.9104 & 92.73\% & 0.9275 \\[2pt]
    & 500 & 99.55\% & 0.9955 & 98.64\% & 0.9864 & 93.86\% & 0.9381 & 92.73\% & 0.9256 & 98.86\% & 0.9885 & 99.55\% & 0.9955 & 98.86\% & 0.9885 & 99.09\% & 0.9908 \\[2pt]
    & 1000 & 100.00\% & 1.0000 & 99.32\% & 0.9932 & 99.32\% & 0.9932 & 99.32\% & 0.9932 & 100.00\% & 1.0000 & 100.00\% & 1.0000 & 100.00\% & 1.0000 & 99.55\% & 0.9954\\\bottomrule
    \end{tabular}
\end{table}
\end{landscape}

\begin{landscape}

    \begin{table}
    
    \caption{Experimental results in accuracy and macro-\(F_1\) for tabular datasets, for AML in both classification variants and for the baselines. The headline results for AML use the logistic regression readout; results obtained via fewest misses are also reported.}
    \label{table:results_tabular_datasets}
    \centering
    \begin{tabular}{cccccccccccccccc}
        \toprule
    &\multirow{2}{*}{\# train} & \multicolumn{2}{c}{AML} & \multicolumn{2}{c}{Fewest misses} & \multicolumn{2}{c}{XGBoost} & \multicolumn{2}{c}{LightGBM} & \multicolumn{2}{c}{Random Forest} & \multicolumn{2}{c}{SVM} & \multicolumn{2}{c}{MLP} \\
        && acc.\ & \(F_1\) & acc.\ & \(F_1\) & acc.\ & \(F_1\) & acc.\ & \(F_1\) & acc.\ & \(F_1\)  & acc.\ & \(F_1\) & acc.\ & \(F_1\)\\[2pt] \hline\midrule \\[-7pt]
    
    \multirow{4}{*}{\begin{sideways}\parbox{1.5cm}{ada}\end{sideways}}
            
    & 50 & 71.08\% & 0.6310 & 71.81\% & 0.6433 & 74.94\% & 0.7123 & 66.75\% & 0.6183 & 77.83\% & 0.7304 & 73.25\% & 0.7090 & 79.76\% & 0.7513 \\[2pt]
    & 100 & 75.90\% & 0.7151 & 77.59\% & 0.7254 & 78.80\% & 0.7556 & 75.90\% & 0.7276 & 79.04\% & 0.7333 & 71.57\% & 0.6542 & 75.90\% & 0.7084 \\[2pt]
    & 200 & 77.11\% & 0.7054 & 78.55\% & 0.7154 & 74.94\% & 0.7253 & 71.08\% & 0.6919 & 79.28\% & 0.7454 & 71.81\% & 0.6160 & 73.98\% & 0.6865 \\[2pt]
    & 500 & 82.17\% & 0.7725 & 82.17\% & 0.7656 & 80.00\% & 0.7630 & 80.96\% & 0.7713 & 82.65\% & 0.7733 & 76.14\% & 0.7298 & 78.80\% & 0.7177 \\[2pt]
    & 1000 & 82.65\% & 0.7747 & 82.89\% & 0.7730 & 78.31\% & 0.7548 & 81.93\% & 0.7877 & 82.65\% & 0.7786 & 80.00\% & 0.7640 & 81.20\% & 0.7691 \\[2pt]
    & 2000 & 83.13\% & 0.7769 & 83.13\% & 0.7710 & 82.65\% & 0.7984 & 81.69\% & 0.7881 & 83.86\% & 0.7969 & 80.00\% & 0.7640 & 81.20\% & 0.7514 \\[2pt]
    
\hline\\[-6pt]

    \multirow{4}{*}{\begin{sideways}\parbox{1.55cm}{Australian}\end{sideways}}

    & 50 & 72.46\% & 0.7244 & 73.91\% & 0.7391 & 88.41\% & 0.8834 & 86.96\% & 0.8678 & 86.96\% & 0.8686 & 88.41\% & 0.8829 & 86.96\% & 0.8686 \\[2pt]
    & 100 & 78.26\% & 0.7826 & 79.71\% & 0.7971 & 84.06\% & 0.8404 & 85.51\% & 0.8550 & 85.51\% & 0.8550 & 84.06\% & 0.8384 & 85.51\% & 0.8543 \\[2pt]
    & 200 & 89.86\% & 0.8986 & 88.41\% & 0.8840 & 89.86\% & 0.8986 & 89.86\% & 0.8985 & 91.30\% & 0.9129 & 86.96\% & 0.8695 & 91.30\% & 0.9130 \\[2pt]
    & 500 & 91.30\% & 0.9130 & 91.30\% & 0.9130 & 86.96\% & 0.8696 & 86.96\% & 0.8695 & 92.75\% & 0.9275 & 86.96\% & 0.8686 & 91.30\% & 0.9130 \\[2pt]
    \hline\\[-6pt]

\multirow{4}{*}{\begin{sideways}\parbox{1.7685cm}{\centering \small blood-transfusion-service-center}\end{sideways}}

    & 50 & 77.33\% & 0.6832 & 76.00\% & 0.6577 & 64.00\% & 0.5817 & 49.33\% & 0.4722 & 66.67\% & 0.6032 & 76.00\% & 0.6828 & 80.00\% & 0.6612 \\[2pt]
    & 100 & 74.67\% & 0.6460 & 74.67\% & 0.6460 & 73.33\% & 0.6345 & 73.33\% & 0.6591 & 73.33\% & 0.6028 & 76.00\% & 0.5813 & 76.00\% & 0.6577 \\[2pt]
    & 200 & 72.00\% & 0.5921 & 72.00\% & 0.5921 & 66.67\% & 0.6032 & 66.67\% & 0.5925 & 73.33\% & 0.6345 & 77.33\% & 0.5920 & 77.33\% & 0.6698 \\[2pt]
    & 500 & 68.00\% & 0.5436 & 68.00\% & 0.5436 & 70.67\% & 0.6462 & 73.33\% & 0.6865 & 74.67\% & 0.6903 & 78.67\% & 0.5385 & 80.00\% & 0.6794 \\[2pt]
    \hline\\[-6pt]

\multirow{4}{*}{\begin{sideways}\parbox{2.1cm}{\centering \small car}\end{sideways}}
    & 50 & 78.03\% & 0.4419 & 79.77\% & 0.4694 & 74.57\% & 0.4573 & 54.34\% & 0.3222 & 77.46\% & 0.4843 & 80.35\% & 0.5689 & 80.35\% & 0.5266 \\[2pt]
    & 100 & 83.82\% & 0.6193 & 83.24\% & 0.6186 & 89.60\% & 0.7884 & 82.08\% & 0.6267 & 86.71\% & 0.6711 & 88.44\% & 0.6980 & 90.17\% & 0.7635 \\[2pt]
    & 200 & 89.02\% & 0.6992 & 88.44\% & 0.6732 & 89.02\% & 0.7568 & 90.17\% & 0.7577 & 92.49\% & 0.7757 & 91.91\% & 0.7725 & 93.06\% & 0.8080 \\[2pt]
    & 500 & 94.22\% & 0.8309 & 91.91\% & 0.6847 & 93.64\% & 0.8283 & 95.95\% & 0.8834 & 94.80\% & 0.8425 & 96.53\% & 0.9156 & 95.38\% & 0.9038 \\[2pt]
    & 1000 & 98.84\% & 0.9332 & 98.84\% & 0.9332 & 96.53\% & 0.8969 & 97.11\% & 0.9014 & 97.69\% & 0.9059 & 97.69\% & 0.9240 & 98.27\% & 0.9203 \\[2pt]
    \hline\\[-6pt]

\multirow{4}{*}{\begin{sideways}\parbox{2.5cm}{\centering \small churn}\end{sideways}}
    & 50 & 83.40\% & 0.5508 & 84.20\% & 0.4921 & 79.40\% & 0.5355 & 69.40\% & 0.5049 & 84.80\% & 0.4837 & 79.20\% & 0.6315 & 84.40\% & 0.4577 \\[2pt]
    & 100 & 84.00\% & 0.6266 & 85.00\% & 0.6010 & 83.00\% & 0.6167 & 80.60\% & 0.5947 & 84.20\% & 0.5650 & 85.20\% & 0.5482 & 84.40\% & 0.4577 \\[2pt]
    & 200 & 86.40\% & 0.6471 & 87.00\% & 0.6356 & 87.40\% & 0.7668 & 84.00\% & 0.6368 & 87.00\% & 0.6599 & 86.80\% & 0.6459 & 86.80\% & 0.6733 \\[2pt]
    & 500 & 88.00\% & 0.7159 & 87.80\% & 0.7048 & 88.20\% & 0.7794 & 83.60\% & 0.7529 & 88.80\% & 0.7185 & 87.00\% & 0.7145 & 87.20\% & 0.7054 \\[2pt]
    & 1000 & 88.00\% & 0.6984 & 88.20\% & 0.6693 & 92.40\% & 0.8476 & 93.80\% & 0.8778 & 92.00\% & 0.8159 & 88.80\% & 0.7587 & 88.40\% & 0.7038 \\[2pt]
    & 2000 & 88.80\% & 0.7140 & 87.00\% & 0.6220 & 93.80\% & 0.8841 & 95.20\% & 0.9038 & 94.80\% & 0.8920 & 91.00\% & 0.8097 & 89.60\% & 0.8174 \\[2pt]
\bottomrule
\end{tabular}
\end{table}

\begin{table}
    \centering
    \begin{tabular}{cccccccccccccccc}
        \toprule
    &\multirow{2}{*}{\# train} & \multicolumn{2}{c}{AML} & \multicolumn{2}{c}{Fewest misses}  &\multicolumn{2}{c}{XGBoost} & \multicolumn{2}{c}{LightGBM} & \multicolumn{2}{c}{Random Forest} & \multicolumn{2}{c}{SVM} & \multicolumn{2}{c}{MLP} \\
        && acc.\ & \(F_1\) & acc.\ & \(F_1\) & acc.\ & \(F_1\) & acc.\ & \(F_1\) & acc.\ & \(F_1\)  & acc.\ & \(F_1\) & acc.\ & \(F_1\)\\[2pt] \hline\midrule \\[-7pt]

\multirow{4}{*}{\begin{sideways}\parbox{2.1cm}{\centering \small cmc}\end{sideways}}

    & 50 & 43.92\% & 0.4342 & 44.59\% & 0.4403 & 47.30\% & 0.4738 & 35.81\% & 0.3516 & 43.92\% & 0.4351 & 42.57\% & 0.4136 & 46.62\% & 0.4403 \\[2pt]
    & 100 & 52.03\% & 0.5181 & 52.70\% & 0.5260 & 47.97\% & 0.4774 & 49.32\% & 0.4924 & 48.65\% & 0.4861 & 47.30\% & 0.4723 & 48.65\% & 0.4842 \\[2pt]
    & 200 & 50.68\% & 0.5074 & 53.38\% & 0.5379 & 47.97\% & 0.4800 & 47.30\% & 0.4748 & 47.30\% & 0.4754 & 54.05\% & 0.5448 & 52.03\% & 0.5213 \\[2pt]
    & 500 & 50.68\% & 0.4963 & 52.70\% & 0.5134 & 50.68\% & 0.5062 & 49.32\% & 0.4918 & 47.30\% & 0.4761 & 44.59\% & 0.4468 & 52.03\% & 0.5043 \\[2pt]
    & 1000 & 56.76\% & 0.5350 & 56.08\% & 0.5321 & 49.32\% & 0.4901 & 46.62\% & 0.4701 & 49.32\% & 0.4960 & 52.03\% & 0.5149 & 52.70\% & 0.5281 \\[2pt]
    \hline\\[-6pt]

\multirow{4}{*}{\begin{sideways}\parbox{1.7cm}{\centering \small credit-g}\end{sideways}}

    & 50 & 60.00\% & 0.4667 & 56.00\% & 0.3789 & 57.00\% & 0.4725 & 58.00\% & 0.5000 & 61.00\% & 0.4729 & 63.00\% & 0.5131 & 60.00\% & 0.4667 \\[2pt]
    & 100 & 65.00\% & 0.5793 & 66.00\% & 0.5784 & 67.00\% & 0.6397 & 62.00\% & 0.6194 & 67.00\% & 0.6349 & 59.00\% & 0.3711 & 72.00\% & 0.6727 \\[2pt]
    & 200 & 70.00\% & 0.6608 & 71.00\% & 0.6640 & 67.00\% & 0.6692 & 70.00\% & 0.6956 & 71.00\% & 0.6792 & 68.00\% & 0.6435 & 71.00\% & 0.6580 \\[2pt]
    & 500 & 71.00\% & 0.6695 & 74.00\% & 0.6905 & 71.00\% & 0.7050 & 65.00\% & 0.6457 & 74.00\% & 0.6843 & 66.00\% & 0.6212 & 71.00\% & 0.6938 \\[2pt]
    \hline\\[-6pt]

\multirow{4}{*}{\begin{sideways}\parbox{2.5cm}{\centering \small dna}\end{sideways}}

    & 50 & 59.25\% & 0.5648 & 57.68\% & 0.5226 & 86.21\% & 0.8527 & 62.38\% & 0.6168 & 84.64\% & 0.8273 & 75.55\% & 0.7285 & 69.59\% & 0.6266 \\[2pt]
    & 100 & 81.19\% & 0.7938 & 80.56\% & 0.7910 & 85.89\% & 0.8534 & 82.45\% & 0.8175 & 84.95\% & 0.8357 & 78.68\% & 0.7634 & 79.94\% & 0.7715 \\[2pt]
    & 200 & 89.03\% & 0.8808 & 85.89\% & 0.8427 & 89.34\% & 0.8898 & 87.15\% & 0.8630 & 88.71\% & 0.8780 & 86.52\% & 0.8552 & 85.27\% & 0.8372 \\[2pt]
    & 500 & 91.85\% & 0.9136 & 91.22\% & 0.9069 & 91.22\% & 0.9086 & 89.34\% & 0.8869 & 91.85\% & 0.9126 & 91.85\% & 0.9131 & 90.91\% & 0.9022 \\[2pt]
    & 1000 & 90.60\% & 0.9001 & 91.54\% & 0.9089 & 91.22\% & 0.9063 & 90.91\% & 0.9042 & 92.48\% & 0.9192 & 91.54\% & 0.9097 & 90.91\% & 0.9008 \\[2pt]
    & 2000 & 92.79\% & 0.9236 & 92.16\% & 0.9164 & 93.42\% & 0.9267 & 93.10\% & 0.9240 & 93.42\% & 0.9255 & 93.10\% & 0.9232 & 92.16\% & 0.9119 \\[2pt]
    \hline\\[-6pt]

\multirow{4}{*}{\begin{sideways}\parbox{1.7cm}{\centering \small eucalyptus}\end{sideways}}

    & 50 & 47.30\% & 0.4563 & 44.59\% & 0.4474 & 40.54\% & 0.4217 & 32.43\% & 0.3010 & 35.14\% & 0.3614 & 41.89\% & 0.4119 & 45.95\% & 0.3859 \\[2pt]
    & 100 & 50.00\% & 0.4677 & 54.05\% & 0.5159 & 51.35\% & 0.5165 & 52.70\% & 0.5092 & 56.76\% & 0.5706 & 58.11\% & 0.5876 & 64.86\% & 0.6331 \\[2pt]
    & 200 & 55.41\% & 0.5362 & 67.57\% & 0.6670 & 58.11\% & 0.5751 & 56.76\% & 0.5617 & 60.81\% & 0.6042 & 62.16\% & 0.6202 & 55.41\% & 0.5482 \\[2pt]
    & 500 & 70.27\% & 0.6953 & 71.62\% & 0.7041 & 71.62\% & 0.7173 & 70.27\% & 0.7022 & 68.92\% & 0.6902 & 62.16\% & 0.6084 & 72.97\% & 0.7139 \\[2pt]
    \hline\\[-6pt]

\multirow{4}{*}{\begin{sideways}\parbox{2.45cm}{\centering \small first-order-theorem-proving}\end{sideways}}

    & 50 & 37.75\% & 0.2181 & 39.22\% & 0.1946 & 38.89\% & 0.2569 & 35.95\% & 0.2644 & 37.91\% & 0.2186 & 41.99\% & 0.1352 & 41.67\% & 0.0980 \\[2pt]
    & 100 & 39.38\% & 0.2740 & 43.30\% & 0.2814 & 37.25\% & 0.2634 & 27.61\% & 0.2390 & 42.32\% & 0.2784 & 41.18\% & 0.1759 & 37.42\% & 0.2571 \\[2pt]
    & 200 & 39.54\% & 0.2878 & 41.01\% & 0.2816 & 38.73\% & 0.3096 & 39.38\% & 0.2960 & 43.79\% & 0.3008 & 41.83\% & 0.2148 & 40.03\% & 0.2738 \\[2pt]
    & 500 & 44.77\% & 0.3462 & 44.93\% & 0.3327 & 40.69\% & 0.3304 & 37.42\% & 0.3151 & 42.32\% & 0.3497 & 39.38\% & 0.2854 & 41.50\% & 0.3019 \\[2pt]
    & 1000 & 53.10\% & 0.4246 & 51.80\% & 0.4018 & 47.39\% & 0.3946 & 39.54\% & 0.3346 & 50.33\% & 0.3959 & 46.24\% & 0.3414 & 47.06\% & 0.3196 \\[2pt]
    & 2000 & 55.39\% & 0.4434 & 56.05\% & 0.4423 & 52.12\% & 0.4353 & 52.94\% & 0.4395 & 54.58\% & 0.4462 & 44.44\% & 0.3582 & 49.51\% & 0.3723 \\[2pt]
    \bottomrule

\end{tabular}
\end{table}

\begin{table}[ht]
    \centering
    \begin{tabular}{cccccccccccccccc}
        \toprule
    &\multirow{2}{*}{\# train} & \multicolumn{2}{c}{AML} & \multicolumn{2}{c}{Fewest misses}  &\multicolumn{2}{c}{XGBoost} & \multicolumn{2}{c}{LightGBM} & \multicolumn{2}{c}{Random Forest} & \multicolumn{2}{c}{SVM} & \multicolumn{2}{c}{MLP} \\
        && acc.\ & \(F_1\) & acc.\ & \(F_1\) & acc.\ & \(F_1\) & acc.\ & \(F_1\) & acc.\ & \(F_1\)  & acc.\ & \(F_1\) & acc.\ & \(F_1\)\\[2pt] \hline\midrule \\[-7pt]

\multirow{4}{*}{\begin{sideways}\parbox{2.5cm}{\centering \small GesturePhase Segmentation Processed}\end{sideways}}

    & 50 & 37.85\% & 0.3079 & 39.17\% & 0.3018 & 36.34\% & 0.3211 & 36.54\% & 0.3217 & 39.98\% & 0.3143 & 37.96\% & 0.3020 & 37.55\% & 0.2654 \\[2pt]
    & 100 & 42.81\% & 0.3429 & 41.50\% & 0.3160 & 38.97\% & 0.3592 & 37.55\% & 0.3533 & 42.61\% & 0.3473 & 41.90\% & 0.2617 & 42.41\% & 0.3219 \\[2pt]
    & 200 & 48.48\% & 0.3676 & 49.29\% & 0.3507 & 47.67\% & 0.4033 & 42.31\% & 0.3733 & 48.89\% & 0.3666 & 40.79\% & 0.3473 & 47.77\% & 0.3607 \\[2pt]
    & 500 & 52.33\% & 0.4344 & 51.82\% & 0.3921 & 48.58\% & 0.4310 & 45.85\% & 0.4195 & 52.33\% & 0.3911 & 47.17\% & 0.3999 & 48.28\% & 0.3132 \\[2pt]
    & 1000 & 54.25\% & 0.4747 & 54.25\% & 0.4413 & 52.43\% & 0.4679 & 48.48\% & 0.4531 & 54.45\% & 0.4292 & 48.79\% & 0.3846 & 50.20\% & 0.4431 \\[2pt]
    & 2000 & 58.30\% & 0.5256 & 56.28\% & 0.4905 & 58.70\% & 0.5382 & 57.69\% & 0.5356 & 57.09\% & 0.4833 & 51.11\% & 0.4323 & 47.17\% & 0.4313 \\[2pt]

\hline\\[-6pt]

\multirow{4}{*}{\begin{sideways}\parbox{2.5cm}{\centering \small jasmine}\end{sideways}}

    & 50 & 69.90\% & 0.6987 & 69.23\% & 0.6919 & 74.92\% & 0.7490 & 76.92\% & 0.7690 & 74.92\% & 0.7492 & 79.93\% & 0.7962 & 73.58\% & 0.7355 \\[2pt]
    & 100 & 69.90\% & 0.6972 & 70.57\% & 0.7039 & 77.93\% & 0.7787 & 70.23\% & 0.7021 & 79.26\% & 0.7894 & 78.60\% & 0.7790 & 76.59\% & 0.7647 \\[2pt]
    & 200 & 76.25\% & 0.7623 & 77.93\% & 0.7788 & 79.93\% & 0.7981 & 80.27\% & 0.8021 & 79.93\% & 0.7962 & 80.94\% & 0.8034 & 73.24\% & 0.7323 \\[2pt]
    & 500 & 78.26\% & 0.7825 & 79.26\% & 0.7918 & 81.94\% & 0.8185 & 81.61\% & 0.8151 & 81.61\% & 0.8144 & 79.93\% & 0.7965 & 77.93\% & 0.7793 \\[2pt]
    & 1000 & 80.94\% & 0.8089 & 80.27\% & 0.8014 & 82.27\% & 0.8218 & 82.27\% & 0.8209 & 81.61\% & 0.8149 & 80.60\% & 0.8007 & 80.60\% & 0.8060 \\[2pt]
    & 2000 & 82.61\% & 0.8259 & 84.28\% & 0.8424 & 84.62\% & 0.8452 & 83.28\% & 0.8318 & 82.61\% & 0.8242 & 80.27\% & 0.8016 & 77.59\% & 0.7759 \\[2pt]
    \hline\\[-6pt]

\multirow{4}{*}{\begin{sideways}\parbox{2.1cm}{\centering \small kc1}\end{sideways}}

& 50 & 84.36\% & 0.6049 & 83.41\% & 0.5468 & 83.89\% & 0.5305 & 82.94\% & 0.5608 & 82.94\% & 0.5433 & 83.89\% & 0.6004 & 82.94\% & 0.4534 \\[2pt]
& 100 & 79.15\% & 0.4418 & 82.46\% & 0.4519 & 78.20\% & 0.5968 & 80.09\% & 0.5802 & 83.41\% & 0.5272 & 83.89\% & 0.5084 & 82.94\% & 0.4534 \\[2pt]
& 200 & 80.09\% & 0.5676 & 81.99\% & 0.5833 & 78.67\% & 0.6742 & 73.46\% & 0.6279 & 80.57\% & 0.5428 & 83.41\% & 0.5056 & 83.41\% & 0.5468 \\[2pt]
& 500 & 81.52\% & 0.6362 & 81.52\% & 0.6045 & 64.93\% & 0.5851 & 69.19\% & 0.6115 & 82.94\% & 0.6052 & 82.94\% & 0.5241 & 83.41\% & 0.6224 \\[2pt]
& 1000 & 82.46\% & 0.6718 & 81.99\% & 0.6088 & 73.93\% & 0.6371 & 68.25\% & 0.6038 & 81.99\% & 0.5833 & 83.41\% & 0.5468 & 82.94\% & 0.4534 \\[2pt]
    \hline\\[-6pt]

\multirow{4}{*}{\begin{sideways}\parbox{2.5cm}{\centering \small kr-vs-kp}\end{sideways}}
    
    & 50 & 87.81\% & 0.8776 & 83.12\% & 0.8288 & 88.12\% & 0.8811 & 59.69\% & 0.5962 & 85.00\% & 0.8500 & 75.00\% & 0.7496 & 51.88\% & 0.3832 \\[2pt]
    & 100 & 84.69\% & 0.8468 & 84.38\% & 0.8437 & 90.31\% & 0.9031 & 85.94\% & 0.8591 & 91.56\% & 0.9156 & 83.12\% & 0.8307 & 87.81\% & 0.8781 \\[2pt]
    & 200 & 95.00\% & 0.9500 & 94.69\% & 0.9469 & 97.19\% & 0.9719 & 95.00\% & 0.9500 & 95.31\% & 0.9531 & 91.88\% & 0.9187 & 90.94\% & 0.9094 \\[2pt]
    & 500 & 97.81\% & 0.9781 & 97.81\% & 0.9781 & 98.44\% & 0.9844 & 98.12\% & 0.9812 & 97.50\% & 0.9750 & 95.31\% & 0.9531 & 96.56\% & 0.9656 \\[2pt]
    & 1000 & 98.75\% & 0.9875 & 98.44\% & 0.9844 & 99.06\% & 0.9906 & 99.69\% & 0.9969 & 98.75\% & 0.9875 & 96.88\% & 0.9687 & 96.88\% & 0.9687 \\[2pt]
    & 2000 & 98.75\% & 0.9875 & 98.75\% & 0.9875 & 99.38\% & 0.9937 & 99.38\% & 0.9937 & 98.44\% & 0.9844 & 98.12\% & 0.9812 & 98.12\% & 0.9812 \\[2pt]

        \bottomrule
    \end{tabular}
\end{table}

\begin{table}[ht]
    \centering
    \begin{tabular}{cccccccccccccccc}
        \toprule
    &\multirow{2}{*}{\# train} & \multicolumn{2}{c}{AML}  & \multicolumn{2}{c}{Fewest misses}  &\multicolumn{2}{c}{XGBoost} & \multicolumn{2}{c}{LightGBM} & \multicolumn{2}{c}{Random Forest} & \multicolumn{2}{c}{SVM} & \multicolumn{2}{c}{MLP} \\
        && acc.\ & \(F_1\) & acc.\ & \(F_1\) & acc.\ & \(F_1\) & acc.\ & \(F_1\) & acc.\ & \(F_1\)  & acc.\ & \(F_1\) & acc.\ & \(F_1\)\\[2pt] \hline\midrule \\[-7pt]

\multirow{4}{*}{\begin{sideways}\parbox{2.5cm}{\centering \small madeline}\end{sideways}}

    & 50 & 50.64\% & 0.5064 & 50.00\% & 0.5000 & 46.82\% & 0.4672 & 43.63\% & 0.4342 & 50.32\% & 0.5030 & 52.87\% & 0.5280 & 54.14\% & 0.5414 \\[2pt]
    & 100 & 46.82\% & 0.4636 & 47.13\% & 0.4576 & 50.96\% & 0.5062 & 47.13\% & 0.4658 & 51.59\% & 0.5033 & 49.36\% & 0.3788 & 48.41\% & 0.4830 \\[2pt]
    & 200 & 61.46\% & 0.6138 & 62.10\% & 0.6209 & 55.41\% & 0.5515 & 55.41\% & 0.5535 & 55.10\% & 0.5489 & 52.87\% & 0.5174 & 49.04\% & 0.4864 \\[2pt]
    & 500 & 57.32\% & 0.5711 & 57.96\% & 0.5779 & 68.79\% & 0.6837 & 65.61\% & 0.6524 & 61.15\% & 0.6063 & 48.73\% & 0.4858 & 53.18\% & 0.5301 \\[2pt]
    & 1000 & 60.83\% & 0.6065 & 60.19\% & 0.5998 & 79.94\% & 0.7962 & 77.07\% & 0.7673 & 69.75\% & 0.6893 & 58.92\% & 0.5890 & 55.73\% & 0.5573 \\[2pt]
    & 2000 & 69.43\% & 0.6933 & 69.11\% & 0.6904 & 87.26\% & 0.8717 & 83.76\% & 0.8367 & 75.48\% & 0.7517 & 57.01\% & 0.5688 & 57.96\% & 0.5785 \\[2pt]
\hline\\[-6pt]

\multirow{4}{*}{\begin{sideways}\parbox{2.2cm}{\centering \small mfeat-factors}\end{sideways}}
    
    & 50 & 81.50\% & 0.7797 & 76.00\% & 0.7416 & 71.00\% & 0.6668 & 77.50\% & 0.7492 & 73.50\% & 0.6864 & 84.00\% & 0.7908 & 83.50\% & 0.8041 \\[2pt]
    & 100 & 91.00\% & 0.9106 & 90.00\% & 0.8993 & 85.50\% & 0.8366 & 86.50\% & 0.8559 & 87.00\% & 0.8618 & 95.00\% & 0.9485 & 92.50\% & 0.9222 \\[2pt]
    & 200 & 96.00\% & 0.9587 & 94.00\% & 0.9380 & 91.50\% & 0.9152 & 91.50\% & 0.9147 & 92.00\% & 0.9214 & 96.50\% & 0.9649 & 94.00\% & 0.9399 \\[2pt]
    & 500 & 96.50\% & 0.9645 & 95.50\% & 0.9545 & 94.50\% & 0.9451 & 95.50\% & 0.9553 & 93.50\% & 0.9352 & 96.00\% & 0.9587 & 95.50\% & 0.9544 \\[2pt]
    & 1000 & 96.50\% & 0.9645 & 95.00\% & 0.9490 & 95.00\% & 0.9497 & 95.50\% & 0.9550 & 95.00\% & 0.9492 & 97.00\% & 0.9692 & 95.50\% & 0.9537 \\[2pt]

    \hline\\[-6pt]

\multirow{4}{*}{\begin{sideways}\parbox{2.5cm}{\centering \small ozone-level-8hr}\end{sideways}}

    & 50 & 93.31\% & 0.6424 & 95.67\% & 0.5658 & 91.34\% & 0.6711 & 91.34\% & 0.6533 & 95.67\% & 0.5658 & 95.28\% & 0.6544 & 95.28\% & 0.6877 \\[2pt]
    & 100 & 92.13\% & 0.6220 & 94.49\% & 0.6357 & 93.31\% & 0.6892 & 92.13\% & 0.6457 & 93.70\% & 0.6199 & 92.13\% & 0.6457 & 93.31\% & 0.6675 \\[2pt]
    & 200 & 94.09\% & 0.5899 & 94.88\% & 0.5535 & 91.34\% & 0.6103 & 88.19\% & 0.6107 & 93.70\% & 0.6199 & 95.28\% & 0.4879 & 93.70\% & 0.6199 \\[2pt]
    & 500 & 94.09\% & 0.6275 & 95.67\% & 0.6653 & 93.31\% & 0.6675 & 93.70\% & 0.6757 & 94.88\% & 0.6044 & 95.67\% & 0.6222 & 93.70\% & 0.6757 \\[2pt]
    & 1000 & 94.49\% & 0.6357 & 95.67\% & 0.6222 & 90.55\% & 0.6587 & 90.94\% & 0.6273 & 93.70\% & 0.6757 & 94.88\% & 0.6771 & 92.52\% & 0.6921 \\[2pt]
    & 2000 & 94.49\% & 0.6357 & 94.88\% & 0.6044 & 90.55\% & 0.6587 & 90.16\% & 0.6528 & 95.67\% & 0.6222 & 94.49\% & 0.5968 & 93.70\% & 0.5836 \\[2pt]
    
    \hline\\[-6pt]

\multirow{4}{*}{\begin{sideways}\parbox{2.1cm}{\centering \small pc4}\end{sideways}}

    & 50 & 83.56\% & 0.5788 & 86.99\% & 0.5844 & 79.45\% & 0.6129 & 71.92\% & 0.5943 & 82.88\% & 0.5214 & 73.97\% & 0.6014 & 90.41\% & 0.7045 \\[2pt]
    & 100 & 84.93\% & 0.6338 & 86.30\% & 0.6490 & 79.45\% & 0.5655 & 79.45\% & 0.6773 & 85.62\% & 0.6211 & 82.19\% & 0.6067 & 90.41\% & 0.4748 \\[2pt]
    & 200 & 86.30\% & 0.6285 & 89.73\% & 0.6719 & 80.14\% & 0.6444 & 78.08\% & 0.6365 & 89.04\% & 0.6622 & 84.25\% & 0.6267 & 92.47\% & 0.7178 \\[2pt]
    & 500 & 89.73\% & 0.6939 & 89.04\% & 0.6368 & 84.93\% & 0.7056 & 86.99\% & 0.7304 & 89.73\% & 0.6939 & 90.41\% & 0.4748 & 88.36\% & 0.6932 \\[2pt]
    & 1000 & 93.15\% & 0.8025 & 93.15\% & 0.7542 & 90.41\% & 0.7782 & 89.04\% & 0.7580 & 90.41\% & 0.7235 & 93.15\% & 0.7542 & 91.10\% & 0.7347 \\[2pt]
    \bottomrule
    \end{tabular}
\end{table}

\begin{table}
    \centering
    \begin{tabular}{cccccccccccccccc}
        \toprule
    &\multirow{2}{*}{\# train} & \multicolumn{2}{c}{AML}  & \multicolumn{2}{c}{Fewest misses} &\multicolumn{2}{c}{XGBoost} & \multicolumn{2}{c}{LightGBM} & \multicolumn{2}{c}{Random Forest} & \multicolumn{2}{c}{SVM} & \multicolumn{2}{c}{MLP} \\
        && acc.\ & \(F_1\) & acc.\ & \(F_1\) & acc.\ & \(F_1\) & acc.\ & \(F_1\) & acc.\ & \(F_1\)  & acc.\ & \(F_1\) & acc.\ & \(F_1\)\\[2pt] \hline\midrule \\[-7pt]

\multirow{4}{*}{\begin{sideways}\parbox{2.4cm}{\centering \small philippine}\end{sideways}}
    & 50 & 66.95\% & 0.6695 & 67.98\% & 0.6798 & 69.52\% & 0.6942 & 68.66\% & 0.6834 & 69.01\% & 0.6896 & 62.50\% & 0.6214 & 67.29\% & 0.6726 \\[2pt]
    & 100 & 68.66\% & 0.6852 & 69.35\% & 0.6920 & 69.35\% & 0.6925 & 67.29\% & 0.6716 & 68.15\% & 0.6799 & 67.12\% & 0.6677 & 65.41\% & 0.6540 \\[2pt]
    & 200 & 66.44\% & 0.6610 & 67.47\% & 0.6716 & 72.26\% & 0.7224 & 73.29\% & 0.7329 & 69.35\% & 0.6931 & 68.66\% & 0.6854 & 66.61\% & 0.6648 \\[2pt]
    & 500 & 67.12\% & 0.6701 & 67.12\% & 0.6688 & 74.49\% & 0.7441 & 72.77\% & 0.7275 & 70.21\% & 0.7005 & 66.95\% & 0.6654 & 63.36\% & 0.6335 \\[2pt]
    & 1000 & 70.38\% & 0.7038 & 70.55\% & 0.7054 & 73.12\% & 0.7312 & 70.38\% & 0.7037 & 71.23\% & 0.7107 & 66.78\% & 0.6665 & 62.50\% & 0.6247 \\[2pt]
    & 2000 & 68.32\% & 0.6830 & 69.86\% & 0.6982 & 74.49\% & 0.7448 & 73.97\% & 0.7397 & 72.60\% & 0.7258 & 66.10\% & 0.6610 & 65.41\% & 0.6536 \\[2pt]
    \hline\\[-6pt]

\multirow{4}{*}{\begin{sideways}\parbox{2.5cm}{\centering \small phoneme}\end{sideways}}

    & 50 & 76.71\% & 0.6921 & 76.52\% & 0.6858 & 72.83\% & 0.6901 & 78.00\% & 0.7261 & 77.45\% & 0.7114 & 77.82\% & 0.7186 & 77.45\% & 0.6895 \\[2pt]
    & 100 & 79.30\% & 0.7289 & 78.93\% & 0.7159 & 81.89\% & 0.7749 & 76.89\% & 0.7324 & 82.62\% & 0.7786 & 78.19\% & 0.7059 & 80.22\% & 0.7177 \\[2pt]
    & 200 & 81.15\% & 0.7597 & 80.22\% & 0.7440 & 77.45\% & 0.7454 & 80.04\% & 0.7661 & 81.15\% & 0.7628 & 80.41\% & 0.7640 & 78.19\% & 0.7508 \\[2pt]
    & 500 & 83.18\% & 0.7906 & 83.73\% & 0.7954 & 83.73\% & 0.8075 & 83.36\% & 0.8017 & 83.18\% & 0.7923 & 81.52\% & 0.7820 & 81.15\% & 0.7769 \\[2pt]
    & 1000 & 83.73\% & 0.7945 & 84.10\% & 0.7983 & 83.55\% & 0.8070 & 84.84\% & 0.8187 & 82.99\% & 0.7870 & 81.70\% & 0.7838 & 82.26\% & 0.7654 \\[2pt]
    & 2000 & 87.06\% & 0.8392 & 87.99\% & 0.8504 & 87.43\% & 0.8480 & 86.69\% & 0.8397 & 89.09\% & 0.8642 & 84.84\% & 0.8060 & 82.99\% & 0.7980 \\[2pt]
    \hline\\[-6pt]

\multirow{4}{*}{\begin{sideways}\parbox{1.7cm}{\centering \small qsar-biodeg}\end{sideways}}

    & 50 & 86.79\% & 0.8484 & 83.02\% & 0.7948 & 81.13\% & 0.7972 & 76.42\% & 0.7478 & 83.96\% & 0.8114 & 74.53\% & 0.6309 & 86.79\% & 0.8484 \\[2pt]
    & 100 & 82.08\% & 0.8063 & 83.96\% & 0.8174 & 85.85\% & 0.8453 & 82.08\% & 0.8063 & 83.02\% & 0.8154 & 85.85\% & 0.8453 & 91.51\% & 0.9018 \\[2pt]
    & 200 & 86.79\% & 0.8528 & 88.68\% & 0.8720 & 85.85\% & 0.8453 & 83.96\% & 0.8246 & 86.79\% & 0.8528 & 90.57\% & 0.8917 & 94.34\% & 0.9329 \\[2pt]
    & 500 & 87.74\% & 0.8582 & 86.79\% & 0.8460 & 87.74\% & 0.8603 & 83.96\% & 0.8224 & 88.68\% & 0.8657 & 90.57\% & 0.8900 & 92.45\% & 0.9120 \\[2pt]
    \hline\\[-6pt]

\multirow{4}{*}{\begin{sideways}\parbox{2.4cm}{\centering \small Satellite}\end{sideways}}

    & 50 & 99.22\% & 0.6647 & 99.41\% & 0.7842 & 98.04\% & 0.6379 & 98.43\% & 0.5960 & 99.02\% & 0.6404 & 99.02\% & 0.7702 & 99.22\% & 0.4980 \\[2pt]
    & 100 & 99.22\% & 0.6647 & 99.41\% & 0.6985 & 98.82\% & 0.6970 & 98.82\% & 0.6220 & 99.22\% & 0.6647 & 99.22\% & 0.7480 & 99.22\% & 0.4980 \\[2pt]
    & 200 & 99.22\% & 0.6647 & 99.22\% & 0.6647 & 99.22\% & 0.7480 & 99.22\% & 0.7480 & 99.02\% & 0.6404 & 98.82\% & 0.6220 & 99.02\% & 0.4975 \\[2pt]
    & 500 & 99.22\% & 0.6647 & 99.41\% & 0.6985 & 99.02\% & 0.6404 & 98.82\% & 0.6220 & 99.02\% & 0.6404 & 99.22\% & 0.6647 & 99.02\% & 0.4975 \\[2pt]
    & 1000 & 98.82\% & 0.6220 & 99.02\% & 0.6404 & 98.82\% & 0.6220 & 97.25\% & 0.6430 & 99.02\% & 0.6404 & 99.02\% & 0.4975 & 99.02\% & 0.6404 \\[2pt]
    & 2000 & 99.22\% & 0.7480 & 99.22\% & 0.7480 & 99.02\% & 0.7197 & 97.45\% & 0.6112 & 99.02\% & 0.6404 & 99.02\% & 0.7197 & 99.22\% & 0.7480 \\[2pt]
    \bottomrule

    \end{tabular}
\end{table}

\begin{table}[ht]
    \centering
    \begin{tabular}{cccccccccccccccccc}
        \toprule
    &\multirow{2}{*}{\# train} & \multicolumn{2}{c}{AML} & \multicolumn{2}{c}{Fewest misses} &\multicolumn{2}{c}{XGBoost} & \multicolumn{2}{c}{LightGBM} & \multicolumn{2}{c}{Random Forest} & \multicolumn{2}{c}{SVM} & \multicolumn{2}{c}{MLP} \\
        && acc.\ & \(F_1\) & acc.\ & \(F_1\) & acc.\ & \(F_1\) & acc.\ & \(F_1\) & acc.\ & \(F_1\)  & acc.\ & \(F_1\) & acc.\ & \(F_1\)\\[2pt] \hline\midrule \\[-7pt]

\multirow{4}{*}{\begin{sideways}\parbox{2.4cm}{\centering \small segment}\end{sideways}}

    & 50 & 76.62\% & 0.7591 & 71.00\% & 0.6959 & 78.79\% & 0.7911 & 80.52\% & 0.8064 & 82.25\% & 0.8275 & 83.55\% & 0.8371 & 77.92\% & 0.7772 \\[2pt]
    & 100 & 89.18\% & 0.8956 & 87.45\% & 0.8813 & 87.01\% & 0.8780 & 90.91\% & 0.9155 & 87.45\% & 0.8828 & 88.31\% & 0.8881 & 87.45\% & 0.8788 \\[2pt]
    & 200 & 91.77\% & 0.9212 & 90.48\% & 0.9095 & 89.18\% & 0.8970 & 90.48\% & 0.9090 & 91.77\% & 0.9226 & 90.48\% & 0.9108 & 90.04\% & 0.9031 \\[2pt]
    & 500 & 95.24\% & 0.9545 & 95.67\% & 0.9592 & 94.81\% & 0.9511 & 93.94\% & 0.9439 & 93.51\% & 0.9378 & 92.21\% & 0.9239 & 94.81\% & 0.9511 \\[2pt]
    & 1000 & 96.97\% & 0.9708 & 96.97\% & 0.9714 & 95.24\% & 0.9550 & 95.67\% & 0.9592 & 94.81\% & 0.9506 & 93.94\% & 0.9417 & 94.37\% & 0.9455 \\[2pt]
    & 2000 & 97.40\% & 0.9749 & 97.84\% & 0.9789 & 97.40\% & 0.9755 & 98.27\% & 0.9835 & 97.40\% & 0.9755 & 95.67\% & 0.9597 & 95.24\% & 0.9565 \\[2pt]
    
    \hline\\[-6pt]

\multirow{4}{*}{\begin{sideways}\parbox{2.2cm}{\centering \small steel-plates-fault}\end{sideways}}

    & 50 & 58.97\% & 0.4691 & 10.77\% & 0.0810 & 57.95\% & 0.4917 & 56.41\% & 0.5092 & 61.54\% & 0.6180 & 60.00\% & 0.5585 & 61.03\% & 0.4653 \\[2pt]
    & 100 & 63.08\% & 0.6298 & 62.56\% & 0.5343 & 63.59\% & 0.5936 & 61.54\% & 0.6413 & 70.26\% & 0.7413 & 66.67\% & 0.6909 & 64.62\% & 0.6601 \\[2pt]
    & 200 & 73.33\% & 0.7446 & 70.77\% & 0.7460 & 69.74\% & 0.7111 & 67.18\% & 0.6940 & 70.77\% & 0.7430 & 73.33\% & 0.7704 & 70.77\% & 0.6889 \\[2pt]
    & 500 & 76.92\% & 0.8119 & 75.38\% & 0.7652 & 75.90\% & 0.7662 & 79.49\% & 0.8316 & 74.87\% & 0.8124 & 72.82\% & 0.7538 & 71.79\% & 0.7344 \\[2pt]
    & 1000 & 81.54\% & 0.8504 & 77.44\% & 0.7919 & 78.46\% & 0.8250 & 77.95\% & 0.8168 & 75.38\% & 0.7658 & 76.92\% & 0.7788 & 71.28\% & 0.7227 \\[2pt]
    \hline\\[-6pt]

\multirow{4}{*}{\begin{sideways}\parbox{2.4cm}{\centering \small sylvine}\end{sideways}}

    & 50 & 89.47\% & 0.8946 & 89.86\% & 0.8986 & 90.84\% & 0.9084 & 90.45\% & 0.9045 & 88.11\% & 0.8806 & 82.46\% & 0.8240 & 70.57\% & 0.6834 \\[2pt]
    & 100 & 87.52\% & 0.8752 & 88.89\% & 0.8888 & 90.45\% & 0.9045 & 88.50\% & 0.8847 & 90.25\% & 0.9024 & 87.72\% & 0.8770 & 83.63\% & 0.8359 \\[2pt]
    & 200 & 91.03\% & 0.9103 & 90.84\% & 0.9083 & 91.81\% & 0.9181 & 92.40\% & 0.9240 & 91.81\% & 0.9181 & 88.69\% & 0.8868 & 85.96\% & 0.8596 \\[2pt]
    & 500 & 93.37\% & 0.9337 & 92.79\% & 0.9279 & 91.81\% & 0.9181 & 91.03\% & 0.9103 & 92.59\% & 0.9259 & 90.25\% & 0.9025 & 82.85\% & 0.8258 \\[2pt]
    & 1000 & 92.79\% & 0.9279 & 93.18\% & 0.9318 & 93.76\% & 0.9376 & 93.76\% & 0.9376 & 92.98\% & 0.9298 & 90.64\% & 0.9064 & 89.47\% & 0.8944 \\[2pt]
    & 2000 & 93.37\% & 0.9337 & 93.76\% & 0.9376 & 94.93\% & 0.9493 & 95.52\% & 0.9552 & 93.18\% & 0.9318 & 91.62\% & 0.9162 & 89.08\% & 0.8907 \\[2pt]\hline\\[-6pt]
    
\multirow{4}{*}{\begin{sideways}\parbox{1.7cm}{\centering \small vehicle}\end{sideways}}

    & 50 & 61.18\% & 0.5886 & 60.00\% & 0.5742 & 65.88\% & 0.6376 & 58.82\% & 0.5487 & 64.71\% & 0.6198 & 67.06\% & 0.6573 & 62.35\% & 0.5535 \\[2pt]
    & 100 & 65.88\% & 0.6412 & 64.71\% & 0.6255 & 64.71\% & 0.6226 & 63.53\% & 0.6175 & 67.06\% & 0.6338 & 72.94\% & 0.7165 & 71.76\% & 0.7184 \\[2pt]
    & 200 & 72.94\% & 0.7211 & 71.76\% & 0.7066 & 75.29\% & 0.7407 & 77.65\% & 0.7609 & 68.24\% & 0.6502 & 65.88\% & 0.6525 & 74.12\% & 0.7290 \\[2pt]
    & 500 & 76.47\% & 0.7574 & 74.12\% & 0.7285 & 70.59\% & 0.6792 & 74.12\% & 0.7338 & 74.12\% & 0.7150 & 88.24\% & 0.8773 & 82.35\% & 0.8098 \\[2pt]

        \bottomrule
    \end{tabular}
\end{table}

\begin{table}[ht]
    \centering
    \begin{tabular}{cccccccccccccccc}
        \toprule
    &\multirow{2}{*}{\# train} & \multicolumn{2}{c}{AML}  & \multicolumn{2}{c}{Fewest misses} &\multicolumn{2}{c}{XGBoost} & \multicolumn{2}{c}{LightGBM} & \multicolumn{2}{c}{Random Forest} & \multicolumn{2}{c}{SVM} & \multicolumn{2}{c}{MLP} \\
        && acc.\ & \(F_1\) & acc.\ & \(F_1\) & acc.\ & \(F_1\) & acc.\ & \(F_1\) & acc.\ & \(F_1\)  & acc.\ & \(F_1\) & acc.\ & \(F_1\)\\[2pt] \hline\midrule \\[-7pt]

\multirow{4}{*}{\begin{sideways}\parbox{2.5cm}{\centering \small wilt}\end{sideways}}
    & 50 & 50.21\% & 0.3552 & 92.56\% & 0.5306 & 96.28\% & 0.7535 & 87.60\% & 0.5817 & 95.25\% & 0.5278 & 5.17\% & 0.0495 & 95.04\% & 0.4873 \\[2pt]
    & 100 & 89.67\% & 0.6150 & 95.25\% & 0.5278 & 95.87\% & 0.7808 & 95.87\% & 0.7968 & 95.87\% & 0.7511 & 94.83\% & 0.7878 & 95.04\% & 0.4873 \\[2pt]
    & 200 & 96.28\% & 0.7946 & 96.69\% & 0.7809 & 94.01\% & 0.7383 & 92.15\% & 0.7149 & 96.28\% & 0.7857 & 98.14\% & 0.8994 & 95.45\% & 0.5652 \\[2pt]
    & 500 & 97.93\% & 0.8859 & 98.14\% & 0.8905 & 97.52\% & 0.8735 & 97.31\% & 0.8546 & 96.28\% & 0.7857 & 97.93\% & 0.9019 & 95.04\% & 0.4873 \\[2pt]
    & 1000 & 97.93\% & 0.8859 & 97.93\% & 0.8810 & 97.11\% & 0.8577 & 97.93\% & 0.8946 & 97.11\% & 0.8083 & 97.73\% & 0.8902 & 97.93\% & 0.8810 \\[2pt]
    & 2000 & 97.73\% & 0.8770 & 97.93\% & 0.8810 & 98.55\% & 0.9301 & 98.35\% & 0.9156 & 98.76\% & 0.9315 & 97.93\% & 0.9019 & 98.14\% & 0.9069 \\[2pt]
    \hline\\[-6pt]

\multirow{4}{*}{\begin{sideways}\parbox{2.5cm}{\centering \small wine-quality-white}\end{sideways}}

    & 50 & 36.33\% & 0.2313 & 34.69\% & 0.2075 & 37.76\% & 0.2780 & 35.10\% & 0.2321 & 41.63\% & 0.2136 & 18.16\% & 0.0599 & 25.92\% & 0.1342 \\[2pt]
    & 100 & 39.39\% & 0.2258 & 42.24\% & 0.2151 & 40.41\% & 0.2528 & 33.27\% & 0.2211 & 42.65\% & 0.2321 & 43.67\% & 0.1403 & 45.71\% & 0.1769 \\[2pt]
    & 200 & 42.86\% & 0.2507 & 45.10\% & 0.2378 & 40.20\% & 0.2764 & 38.16\% & 0.2605 & 51.02\% & 0.2404 & 44.90\% & 0.1469 & 48.16\% & 0.1640 \\[2pt]
    & 500 & 46.12\% & 0.2714 & 47.96\% & 0.2594 & 47.55\% & 0.3372 & 44.90\% & 0.4195 & 50.20\% & 0.2673 & 48.98\% & 0.2277 & 47.55\% & 0.2504 \\[2pt]
    & 1000 & 52.45\% & 0.3384 & 53.88\% & 0.3509 & 51.22\% & 0.3810 & 47.35\% & 0.3491 & 56.12\% & 0.3450 & 52.65\% & 0.3410 & 46.53\% & 0.1691 \\[2pt]
    & 2000 & 57.14\% & 0.3834 & 59.18\% & 0.4091 & 52.04\% & 0.3801 & 53.06\% & 0.3931 & 61.02\% & 0.3950 & 51.22\% & 0.2864 & 52.04\% & 0.2740 \\[2pt]
    
    \hline\\[-6pt]

\multirow{4}{*}{\begin{sideways}\parbox{2.1cm}{\centering \small yeast}\end{sideways}}

    & 50 & 39.60\% & 0.1924 & 17.45\% & 0.1027 & 36.24\% & 0.2621 & 27.52\% & 0.1985 & 43.62\% & 0.3064 & 40.94\% & 0.3492 & 44.97\% & 0.3105 \\[2pt]
    & 100 & 46.31\% & 0.3386 & 47.65\% & 0.3341 & 48.99\% & 0.4651 & 51.68\% & 0.4557 & 54.36\% & 0.4871 & 48.99\% & 0.4255 & 54.36\% & 0.3936 \\[2pt]
    & 200 & 44.97\% & 0.2981 & 46.31\% & 0.2906 & 48.99\% & 0.3449 & 44.97\% & 0.3284 & 53.02\% & 0.4189 & 51.01\% & 0.4266 & 46.98\% & 0.2172 \\[2pt]
    & 500 & 51.68\% & 0.4794 & 53.02\% & 0.4732 & 54.36\% & 0.5683 & 54.36\% & 0.4592 & 58.39\% & 0.4775 & 51.01\% & 0.4364 & 51.01\% & 0.4995 \\[2pt]
    & 1000 & 54.36\% & 0.5035 & 54.36\% & 0.4820 & 58.39\% & 0.5996 & 57.05\% & 0.4516 & 60.40\% & 0.5878 & 57.72\% & 0.5301 & 53.02\% & 0.5807 \\[2pt]
    \bottomrule

\end{tabular}
\end{table}

\end{landscape}

\end{document}